\documentclass[lettersize,journal]{IEEEtran}
\usepackage{graphicx}
\usepackage{tabularx}
\usepackage{booktabs}
\usepackage{multirow}
\usepackage{makecell}
\usepackage{color}
\usepackage{bm}
\usepackage{amsmath}
\usepackage{breqn}
\usepackage{mathtools}
\usepackage{amsthm}
\usepackage{enumerate}
\usepackage{mathptmx}
\usepackage{mdwmath}
\usepackage{mdwtab}
\usepackage{eqparbox}
\usepackage{eucal}
\usepackage{cuted}
\usepackage{amsfonts}
\usepackage[linesnumbered,ruled,vlined]{algorithm2e}
\usepackage{array}
\usepackage{textcomp}
\usepackage{stfloats}
\usepackage{url}
\usepackage{verbatim}
\usepackage{graphicx}
\usepackage{cite}
\usepackage{float}
\usepackage{enumerate}
\usepackage{tikzsymbols}
\usepackage{algpseudocode}
\usepackage[caption=false,font=footnotesize
,labelfont=sf]{subfig}
\begin{document}
\newpage
\thispagestyle{empty}
\mbox{}
\begin{center}
\Large\textbf{Declaration}
\end{center}

\large This work has been submitted to the IEEE for possible publication. Copyright may be transferred without notice, after which this version may no longer be accessible.

\newpage
\title{ElasticROS: An Elastically Collaborative Robot Operation System for Fog and Cloud Robotics}
\author{
\IEEEauthorblockN{Boyi Liu}
\thanks{Manuscript received June 1, 2022; revised August 26, 2022; accept August 26, 2022. }
\thanks{Boyi Liu, Lujia Wang are with Cheng Kar-Shun Robotics Institute, The Hong Kong University of Science and Technology.\\ email: bliubd@connect.ust.hk, eewanglj@ust.hk.}
\thanks{Ming Liu is with the Thrust of Robotics \& Autonomous Systems, The Hong Kong University of Science and Technology (Guangzhou),Guangzhou, 511400, Guangdong. He is also with the Department of Electronic and Computer Engineering, The Hong Kong University of Science and Technology, Hong Kong. email: eelium@ust.hk.}
}
\markboth{Journal of \LaTeX\ Class Files,~Vol.~X, No.~X, August~2022}%
{Shell \MakeLowercase{\textit{et al.}}: Bare Demo of IEEEtran.cls for IEEE Transactions on Magnetics Journals}

\maketitle
\IEEEtitleabstractindextext{%
\begin{abstract}
Robots are integrating more huge-size models to enrich functions and improve accuracy, which leads to out-of-control computing pressure. And thus robots are encountering bottlenecks in computing power and battery capacity. Fog or cloud robotics is one of the most anticipated theories to address these issues. Approaches of cloud robotics have developed from system-level to node-level. However, the present node-level systems are not flexible enough to dynamically adapt to changing conditions. To address this, we present ElasticROS, which evolves the present node-level systems into an algorithm-level one. ElasticROS is based on ROS and ROS2. For fog and cloud robotics, it is the first robot operating system with algorithm-level collaborative computing. ElasticROS develops elastic collaborative computing to achieve adaptability to dynamic conditions. The collaborative computing algorithm is the core and challenge of ElasticROS. We abstract the problem and then propose an algorithm named ElasAction to address. It is a dynamic action decision algorithm based on online learning, which determines how robots and servers cooperate. The algorithm dynamically updates parameters to adapt to changes of conditions where the robot is currently in. It achieves elastically distributing of computing tasks to robots and servers according to configurations.  In addition, we prove that the regret upper bound of the ElasAction is sublinear, which guarantees its convergence and thus enables ElasticROS to be stable in its elasticity. Finally, we conducted experiments with ElasticROS on common tasks of robotics, including SLAM, grasping and human-robot dialogue, and then measured its performances in latency, CPU usage and power consumption. The algorithm-level ElasticROS performs significantly better than the present node-level system.
\end{abstract}
}
\IEEEdisplaynontitleabstractindextext
\IEEEpeerreviewmaketitle
\begin{figure}[!ht]
\centering
	\subfloat[Core elements of FogROS developed by UC-Berkeley, the present node-level fog and cloud robotic systems. \cite{chenFogROSAdaptiveFramework2021}\cite{ichnowskiFogROSAdaptiveExtensible2022a}.]{\includegraphics[width = 0.5\textwidth]{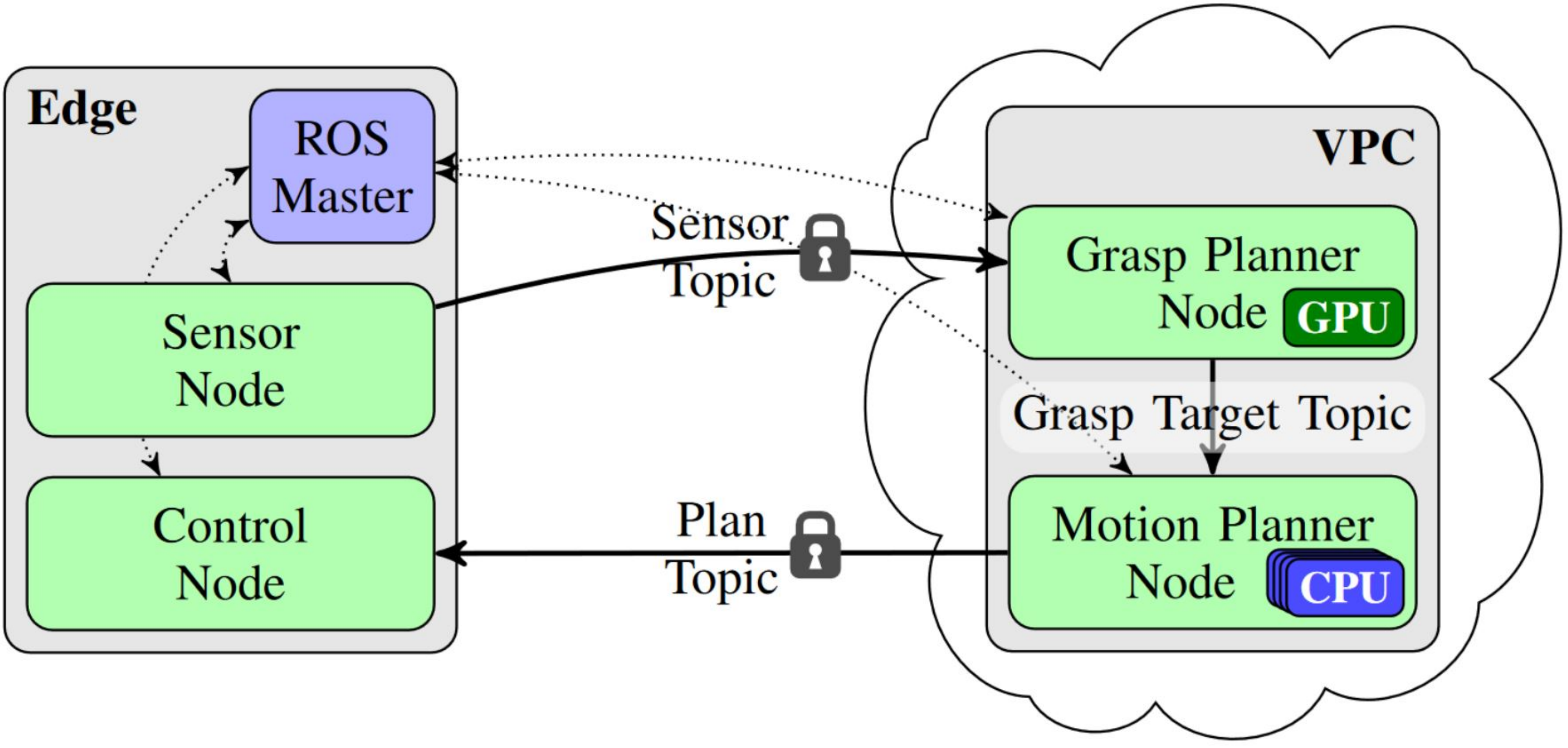}}
	\hfill
	\subfloat[Core elements of ElasticROS. The proposed algorithm-level elastically collaborative computing ROS.]{\includegraphics[width = 0.5\textwidth]{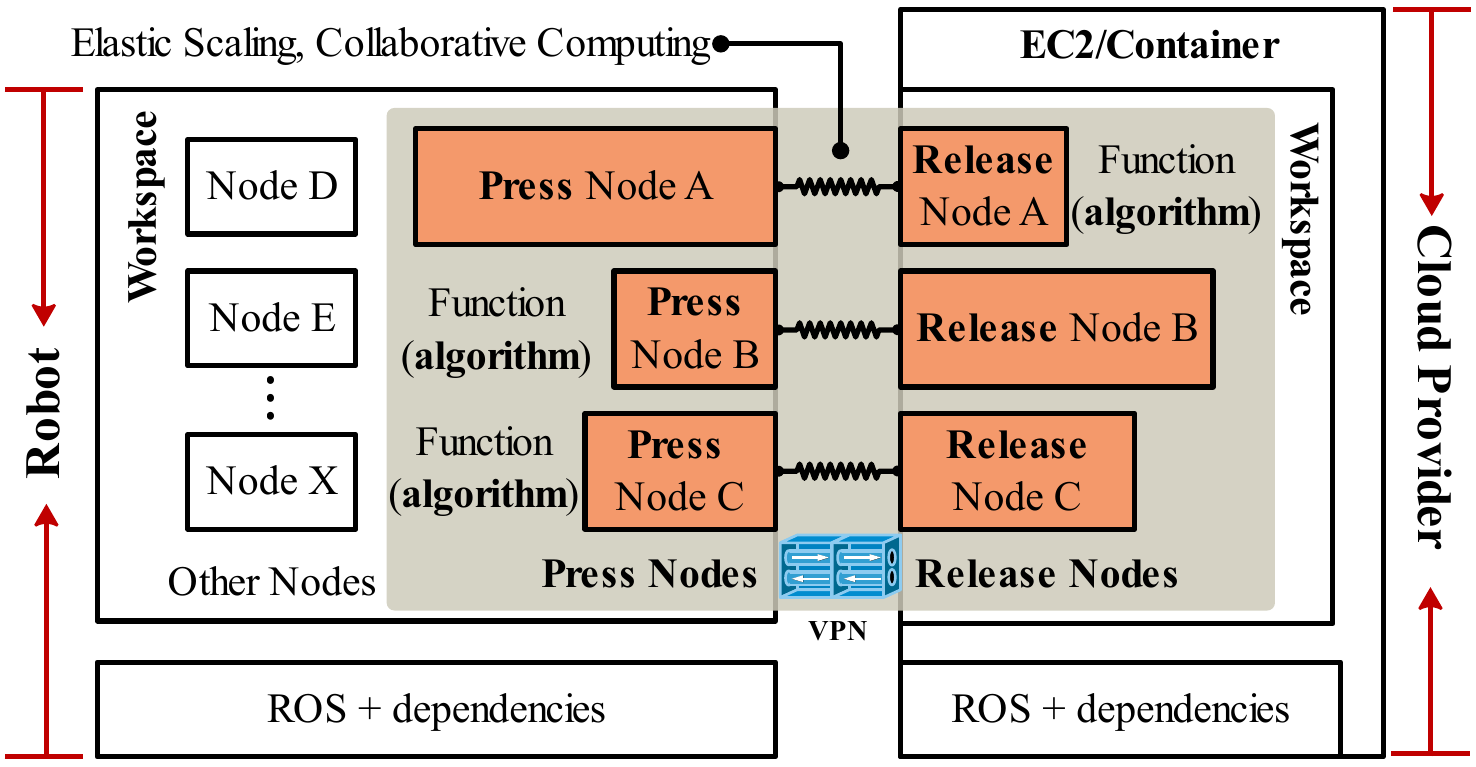}}
\caption{Core elements comparison between the present FogROS and the proposed ElasticROS. The work evolves the present node-level system into an algorithm-level system. ElasticROS is the first robot operating system with algorithm-level collaborative computing for fog or cloud robotics based on ROS and ROS 2.0. Furthermore, the collaborative computing is elastic and dynamic, enabling ElasticROS's self-adaptation to dynamic conditions.}
\label{fig:Firstfig}
\end{figure}
\section{Introduction}
\IEEEPARstart{R}{obots} are integrating more functions with higher computing power demands, as accuracy increases and functionalities become more diverse. Robots are frequently deployed in 3D space, which indicates much larger data processing. The accuracy of perception and control has been greatly improved in recent years. For example, amount of end-to-end machine learning methods with higher power consumption and higher computing complexity are leading in robotic tasks. At the same time, robots are often multi-task integrated agents, taking advantage of all state-of-the-art models or algorithms may result to computation and hardware costs runaway. It is an issue that robots with practical applications (e.g., autonomous vehicles) hard to avoid. Much of the research into accelerating computing in robots has focused on reducing parameters, while the reality is that models are becoming deeper and larger (e.g., deep learning models). Moreover, it has been proved that the depth of the model as DNN is related to the accuracy, and improving the computational speed by cutting parameters will indeed result to a loss in accuracy. Moreover, there exists a limit to the reduction of parameters within the constraints of information theory, so it is not a once-and-for-all way and new approaches should be explored. On the other hand, more mini robots are appearing in the robotics community, and they can hardly afford high computational pressure and power consumptions. However, we are simultaneously trying to enrich and improve their performance on robot, which has caused stumbling blocks in robotics.

Fog or cloud robotics is an anticipated approach to address the above issues. It leverages the cloud or fog server to robotic systems. Cloud robotics is capable of providing functions such as collaborative learning, collaborative localization and mapping, computing offloading, collaborative computing and remote learning, etc.  This previously impossible approach in implementation is becoming a reality with the development of communication technologies (e.g. 5G, 6G). From being proposed with a system-level collaborative way \cite{kuffner2010cloud}, cloud robotics has now evolved to node-level system. The former is data and knowledge collaboration between clouds and robots that have different types of systems. The latter is computational offloading for nodes in a same type of system. Whereas, cloud robotics is still in an initial stage of development and there are still many problems to be overcome. Two most fundamental problems are the theoretical assurance and implementation with a generalized framework based on popular robot operating systems. The two problems are related to the practicality of cloud robotics, which has previously been questioned in the robotics community. In the following, we describe the major challenges about these in detail.

\subsection{Major challenges}
\subsubsection{A generalized system framework in an algorithm-level}Cloud robotics has a variety of applications and lacks a theoretically sound system framework. The present node-level system framework lacks theoretical basis and optimization, and the node-level communication is burdensome. Therefore, there is an urgent requirement for an algorithm-level system framework that optimizes system performance. On the other hand, a number of robotic tasks in various scenarios can be improved by cloud robotics. Each of these tasks has different computing nodes and sensors. A model may work effectively for one robot and undeployable for another. Therefore, it is a challenge to provide a very explicit generalized framework for all robots.
\subsubsection{Dynamic adaptability of the system}Environments where robots are deployed is dynamic, and this is where the difficulty of robotic tasks lies. Cloud robotics are exposed to more dynamic conditions due to the involved network connectivity changes at the same time. Network and computation delays may result to task failures. In addition to the problem of large amounts of transmitted data, the communication of present node-level systems are particularly subject to the problem of communication interruptions. It is not easy to make an optimal collaborative computing decision to improve the system in dynamic conditions. It requires the system to be capable of adapting to a changing conditions and it means a group of dynamic parameters rather than statics, which brings more difficulties to the system design.
\subsubsection{Reliability of the system in theory}It is not easy to make theoretical guarantees on a dynamic robot system. Robots tend to have the implementation of functionality as the main goal, with less attention to the execution of the operating system layer. Therefore, we need to ensure the theoretical reliability of the operating system layer in order to free the user from dependence on the underlying layer. In particular, the communication channel between the server and the robot in a cloud robotic system is uncontrollable. Therefore, theoretical proofs of algorithms related to this channel state are necessary.

\subsection{Contribution}
To address above challenges, this paper contributes in the following ways.
\begin{itemize}
    \item[1)] We present a generalized distributing robot-cloud framework named ElasticROS in an elastically collaborative computing way. It is an algorithm-level framework for fog or cloud robotic system. The system framework of ElasticROS is elegant and advanced, leading to the convenience and robustness in solution. The proposed generalized framework is based on ROS \cite{quigley2009ros} and ROS2 \cite{macenski2022robot}. The two are the most popular systems in the robotics community, which guarantees the generality of ElasticROS. ElasticROS is inspired by cloud computing theories that elastically distribute resources based on various computing tasks, but implement in a more dynamic and complex conditions at an algorithm level.
    \item[2)] We propose ElasticAction, a novel online learning algorithm that enables ElasticROS to compute in an elastically collaborative computing way. ElasticAction develops the computing mechanism of cloud robotics from node-level to algorithm-level. With ElasticAction, ElasticROS is capable of adapting to dynamic conditions in cloud robotic systems. Furthermore, the work abstracts the collaborative computing problem into an end-to-end decision-making problem and addresses it. Users only need to configure the action space and metrics to adopt it.
    \item[3)] In ElasticROS, the ElasticAction in the elastic node controls the elasticity to achieve the user-set metrics optimally. We prove that the regret upper bound of the algorithm is sublinear, which guarantees its convergence and thus enables ElasticROS to be stable in its elasticity.
    \item[4)] We have verified the excellence of ElasticROS with experiments. We experimented the ElasticROS in some computing modules of popular robot tasks such as SLAM, grasping, and human-robot dialogues. We then obtained the experimental results including the metrics of latency, CPU usage, and power consumption, which confirmed the improved performance of ElasticROS.
\end{itemize}
\subsection{Outline}
The rest of this article is organized as follows. Section II consists of the introduction of cloud robotics and related works to ElasticROS. Section III presents of the proposed ElasticROS framework. The ElasticAction algorithm is proposed and proved in Section IV. In Section V, we conduct three common robot experiments with comparisons to verify the efficiency of ElasticROS. Finally, Section V concludes this article.
\subsection{Notations}
 Note that we only use cloud to represent cloud and fog in this paper, since fog computing is often considered an extension and a type of cloud computing. Secondly, ROS stands for ROS and ROS2 where we just want to present the meaning of robot operating systems. ROS means the two if they are not described separately in the context. Some of the figures that appear in this paper are based on existing figures in some other published papers, we inspired by them and drawn new figures according to the ideas of this work. References for figures are \cite{chenFogROSAdaptiveFramework2021,ichnowskiFogROSAdaptiveExtensible2022a,kronauer2021latency}. In addition, all modules related to communication protocols, data compression in previous works are not considered because this is another dimension to improve performance, also it is unfair for the work to consider these as metrics. For communications, our work is focused on the raw data transmissions.

\section{Cloud Robotics and Related Work}
Our work focuses on a generalized framework for cloud robotic systems. In the following, we will first introduce cloud robotics and then discuss related work to ElasticROS. The related work subsection is divided into two sections, system-level applications of cloud robotics and node-level frameworks.
\subsection{Cloud robotics}
Cloud robotics is a technology that applies cloud computing to robotics. The powerful computing and storage capabilities of cloud computing provide robots with a smarter ``brain''. The combination of robotics and cloud computing can enhance the ability of individual robots to perform with more complex functions. For cloud robotics, robots with different capabilities distributed around the world can cooperate and share information resources to accomplish larger and more complex tasks. This will broadly expand the application fields of robots, accelerate and simplify the development process of robotic systems, and reduce constructing costs.

The concept of cloud robotics can be traced back about two decades to the advent of “Networked Robotics” \cite{goldberg2002beyond}. \cite{inaba1997remote} described the advantages of using remote computing for robot control in 1997. In 2001, the IEEE Robotics and Automation Society established the Technical Committee on Networked Robotics \cite{nr.org}. In 2010, James Kuffner first proposed “Cloud Robotics” to describe the increasing number of robotics or automation systems that rely on remote data or code for effective operation \cite{Kuffner}. Since then, various researches of cloud robotics have been developed \cite{kehoe2015survey}.
\begin{figure}
    \centering
    \includegraphics[width=0.48\textwidth]{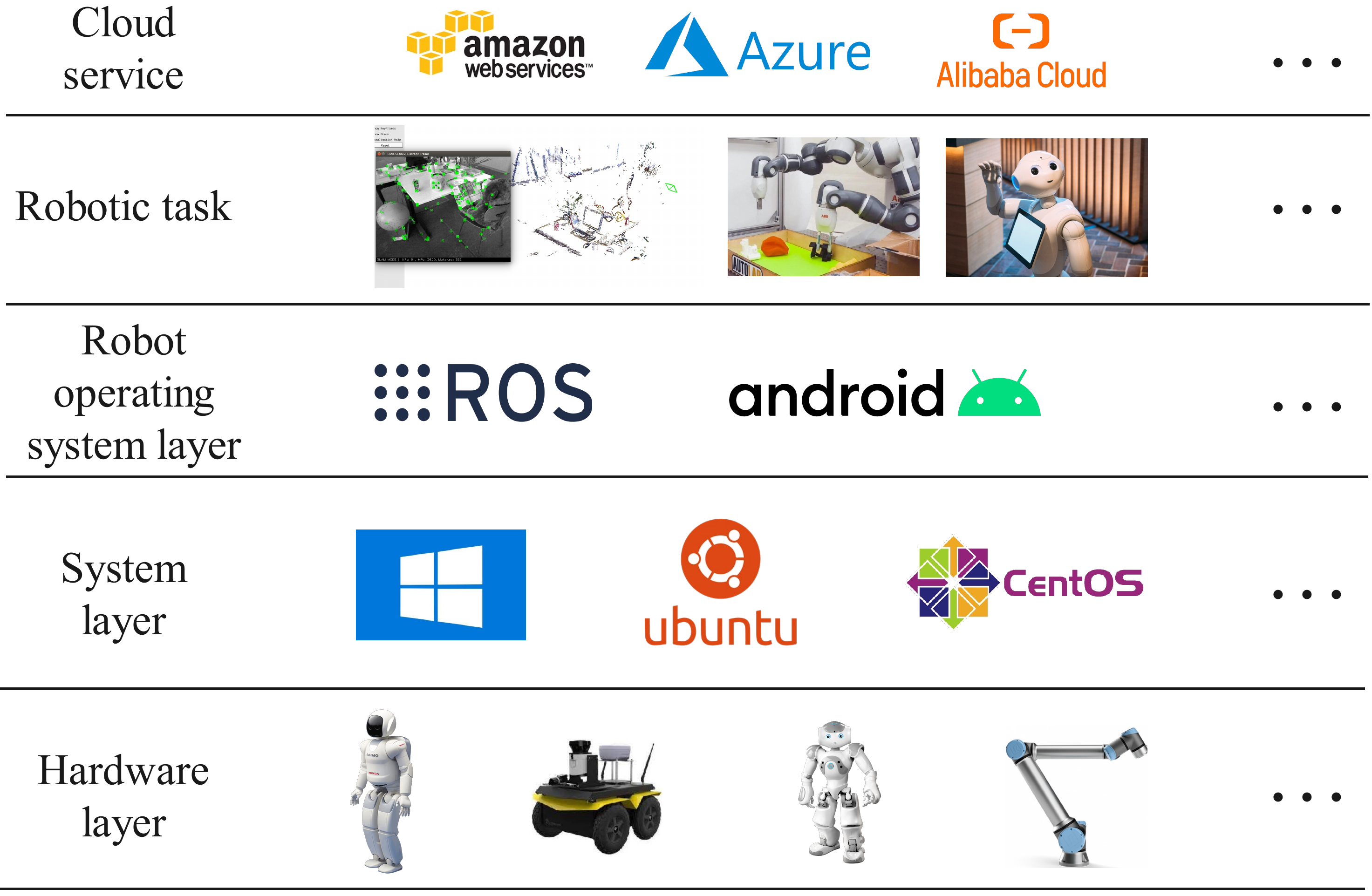}
    \caption{Components of cloud robotics. There are various options for the underlying hardware, systems, robotic operating systems, robotic tasks, and cloud services. The applications of robotics are diverse.}
    \label{fig:my_label}
\end{figure}
\subsection{System-level applications of cloud robotics}
What the system-level means is that the robot and the server are in two separate systems for sharing, offloading and collaboration. System-level applications are the initial form of cloud robotics research to emerge. RoboEarth\cite{waibel2011roboearth} is one of the first cloud robotics applications. It develops the first demonstration of the feasibility of a World Wide Web\cite{riazuelo2015roboearth} for robots. In RoboEarth, robots were able to successfully execute hardware-independent action recipes and could autonomously improve their task performance throughout multiple iterations of execution and knowledge exchange. In the RoboEarth demonstration, four robots collaborate with each other to care for patients in a simulated hospital environment, sharing information and learning from each other by interacting with a cloud-based server. For example, one robot can scan a hospital room and upload the completed map to RoboEarth, while another robot with no knowledge of the room at all can access this map in the cloud to find a glass of water in the room without any additional search. Following the same principle, a similar open-pill-box type of task solving could be shared through RoboEarth, and other robots would not need to be reprogrammed to open a specific box, even if those robots are based on different models. 

For cloud robotic learning systems, \cite{liu2019lifelong} studied the federated learning in the autonomous navigation where the main task is to make the robots fuse and transfer their experience so that they can effectively use prior knowledge and quickly adapt to new environments. They presented the Lifelong Federated Reinforcement Learning (LFRL) and developed a cloud robotic system, in which the robots can learn efficiently in a new environment and extend their experience so that they can use their prior knowledge. Based on this, the authors further propose federated imitation learning \cite{liu2020federated} and peer-assisted learning \cite{liu2021peer} for cloud robotics, which further enables the fusion of heterogeneous data and cloud-based data generation.

\cite{8360045} presents CVI-SLAM, an accurate and powerful system for keyframe-based collaborative SLAM in a type of cloud robotics framework. In CVI-SLAM, participating robots are equipped with a visual-inertial sensor suite and constraint onboard computing power, sharing all information throughout the mission with a more powerful central server. The server merges information from the participating robots and distributes it throughout the system, such that robots can profit from measurements contributed by collaborating robots. \cite{7487342} presents Dex-Net, a dataset of 3D object models and a sampling-based planning algorithm to explore how cloud robotics can be used for robust grasp planning. The algorithm leverages the Google Cloud Platform to simultaneously run up to 1,500 virtual cores, reducing experiment runtime by up to three orders of magnitude.

System-level cloud robotics applications are too diverse for us to list them all. In these applications, the robot and the server are in two separate ROS, or one ROS one another OS, or two separate OSs. In short, the collaboration between robots is built in a primitive way. The heterogeneity of the systems results to difficulties in user deployment and limits the action space available for collaborations. This was the case until node-level frameworks were proposed.
\subsection{Node-level frameworks for cloud robotics}
\begin{figure}
    \centering
    \includegraphics[width=0.5\textwidth]{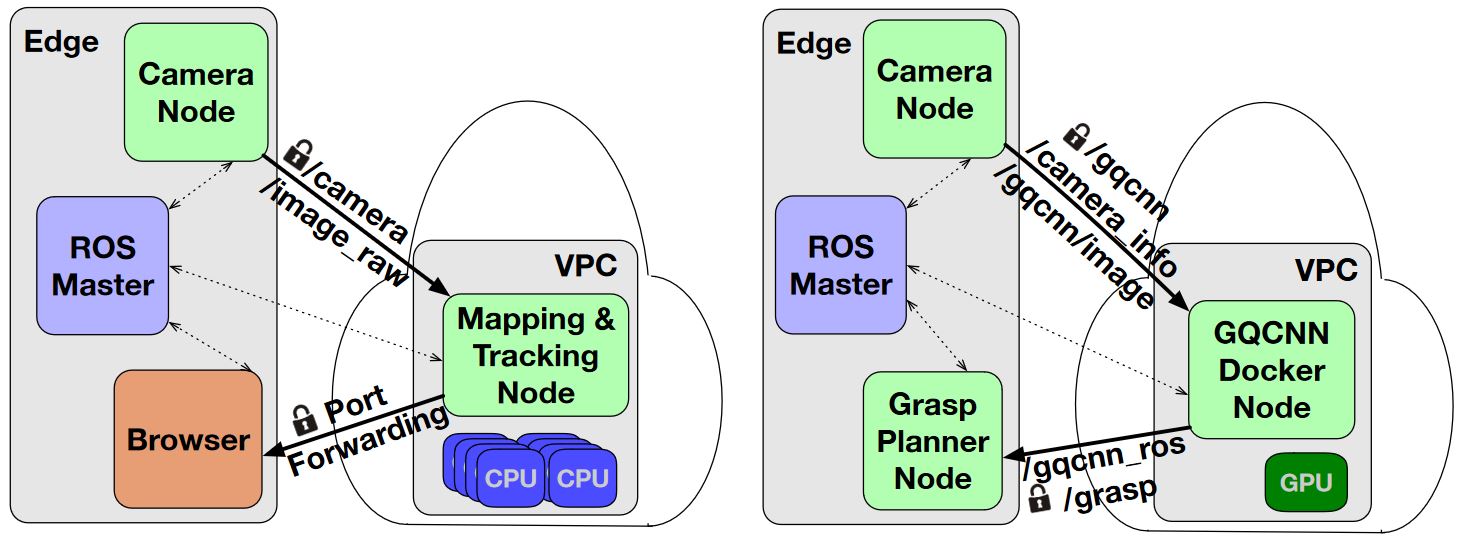}
    \caption{Deployment cases in robotic tasks of VSLAM and Grasping presented in the FogROS paper \cite{chenFogROSAdaptiveFramework2021}. FogROS deploys perception nodes in the robot and computing nodes in the cloud. FogROS is a node-level framework for cloud robotics.}
    \label{fig:FogROSexample}
\end{figure}
There are two branches of FogROS, one ROS based \cite{chenFogROSAdaptiveFramework2021} and one developed for ROS 2 \cite{ichnowskiFogROSAdaptiveExtensible2022a}. They are developed from the same idea, but are implemented in different versions of ROS, so in this article we will use "FogROS" to denote the idea and the corresponding two systems. As shown in Fig. \ref{fig:FogROSexample}, FogROS achieves node-level communication in one system by distributing functional nodes in the robot and the server. FogROS leverages ROS-master as a relay and Data Distribution Service (DDS) in ROS2. In FogROS papers, FogROS performs three robot tasks, SLAM, grasping and path planning. The sensor nodes are deployed on the robot and the computing nodes are deployed in the cloud.  Then the cloud returns the computing results to the robot. FogROS integrates robots and clouds into one ROS, and experimental data demonstrates the improvement of FogROS compared with the system-level computing. However, this node-level framework has inherent drawbacks. Specifically, it has the following problems.
\begin{itemize}
    \item The flexibility of node-level framework is better than that of system-level, but the action space allocated by nodes is still limited and completely depends on the user configuration, without any automatic selection.
    \item The communication of FogROS is heavy, although the authors have made improvements for this by adding some communication protocols \cite{ichnowskiFogROSAdaptiveExtensible2022a}. However, as mentioned in the notation section, data compression should not be taken into account because it is another separately improvement. Especially for sensor nodes, the data they acquired can be very large, in which case FogROS will face a communication crisis.
    \item The computing mode of FogROS is static, but robots work in dynamic conditions, which is a paradox. It means that FogROS cannot response to dynamically changing conditions.
\end{itemize}

In summary, neither the flexibility of the various system-level applications nor the node-level frameworks is up to the task of continuously working under changing conditions. The community expect a more flexible and fine-grained collaborative computing framework with continuous self-adaptive capabilities. ElasticROS fulfills these visions. In the following, we introduce it in detail.
\section{Framework}
In this section, we introduce the framework of ElasticROS, including layer framework and network layout. ElasticROS is implemented based on ROS, but ROS and ROS2 have different framework structures, which leads to different frameworks based on the two. So we introduce ElasticROS's ROS-based and ROS2-based frameworks, respectively.
\subsection{Layer framework of ROS-based ElasticROS}
\begin{figure}
    \centering
        \includegraphics[width=0.49\textwidth]{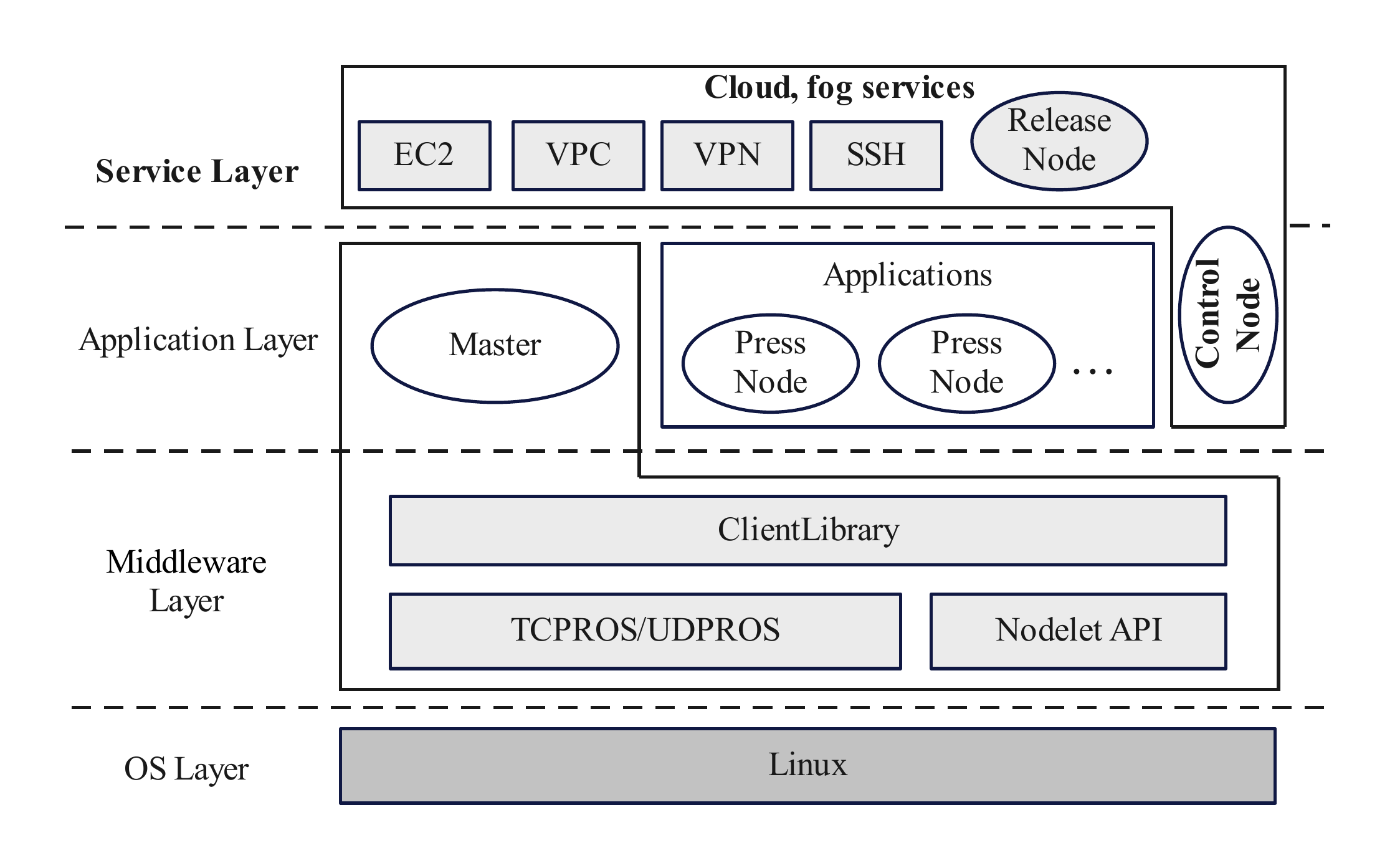}
        \caption{The layer framework of the ROS-based ElasticROS.}
        \label{ROS1}
\end{figure}
 Fig.\ref{ROS1} depicts the layer framework of ElasticROS. ElasticROS is the same as ROS in that it is a Linux-based system framework. It is built on the basic communication protocols, APIs and client libraries in the middleware layer. The layer contains ROS middleware for robot developers, such as the tcpros/udpros communication protocols based on TCP and UDP. The Nodelet for inter-process communication to support real-time data transfer, and a large number of libraries for robot development implementation, such as data type definition, coordinate transformation, and motion control. ElasticROS differs from the ROS layer framework in the service layer and the node layer. Functions in the application layer are implemented with nodes, and ElasticROS develops an elastic control node in the application layer to bridge the service layer and the application layer. At the cloud service layer, there are diverse cloud service providers to select. The release nodes corresponding to the pressure nodes are established in EC2 in the cloud by leveraging tools as VPC, VPN and SSH.

In order to decouple, each function in ROS is a separate process, and each process is running independently. It is a challenge for ElasticROS to build on but break out of this independent functioning mechanism and establish a collaborative computing mechanism. ElasticROS overcomes this challenge by partitioning algorithms to press nodes and release nodes.
\subsection{Network layout of ROS-based ElasticROS}
\begin{figure*}
    \centering
    \includegraphics[width=0.9\textwidth]{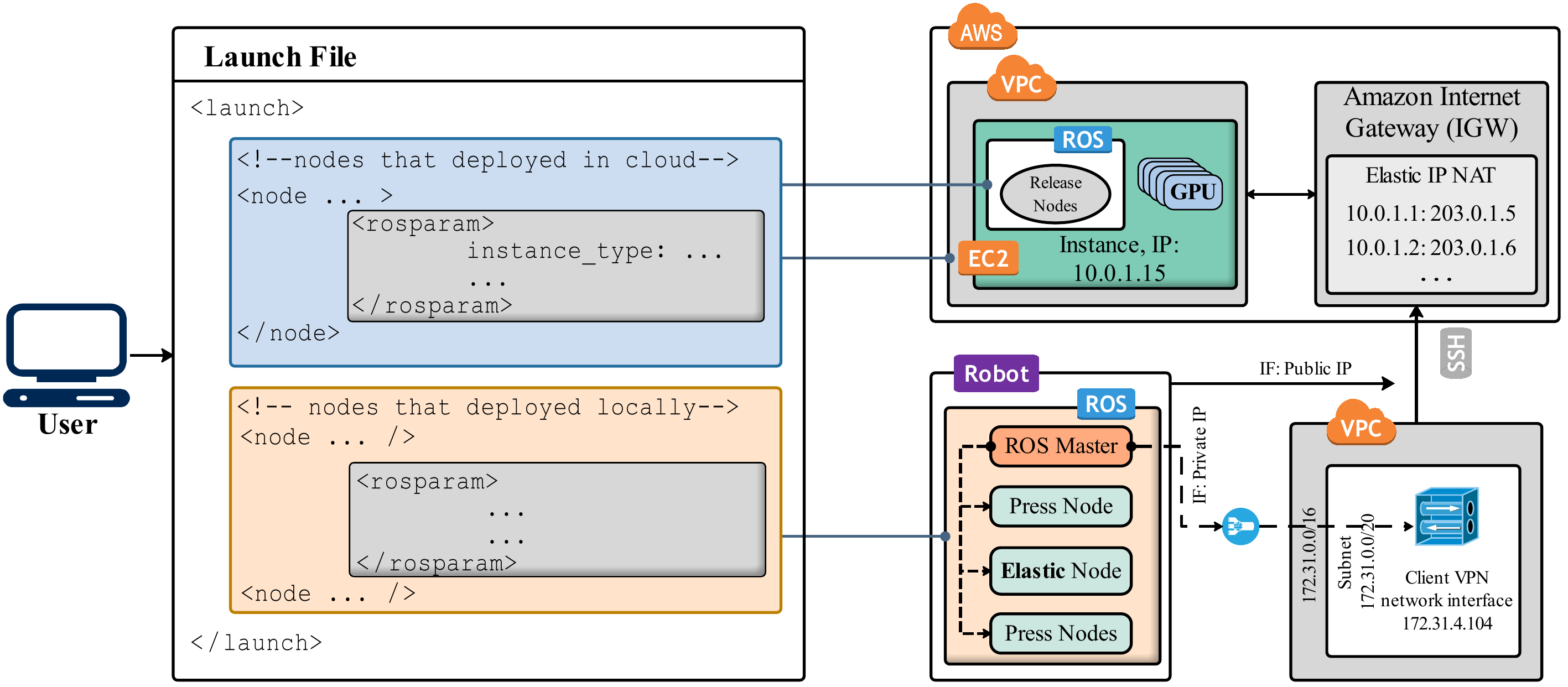}
    \caption{The network layout of ROS-based ElasticROS. ElasticROS reads in the parameters of the Launch file and executes programs after the user has completed the network configuration. It creates SSH, generates nodes, and establishes communication between the cloud and the robot. The press nodes and release nodes are matched according to the action space. Elastic nodes are deployed on the robot for communication decisions.}
    \label{fig:ROS1-framework}
\end{figure*}
As illustrated in Fig.\ref{fig:ROS1-framework}, the ROS-master is responsible for keeping the information registered by Talker and Listener, and matching Talker and Listener with the same topic. Talker and Listener establish a connection with ROS-master, Talker delivers messages, and the messages delivered will be subscribed by Listener. The cloud service launching in ElasticROS is based on FogROS, which launches the application according to the configuration parameters in a Launch file. FogROS has implemented the cloud service easy deployment feature. ElasticROS leverages the module directly but requires additional configuration parameters such as optimization target and action space. The novelty of ElasticROS in this section is the network conversion and the Press-Elastic-Release node format. For the network conversion, this problem existed in the previous version of FogROS \cite{chenFogROSAdaptiveFramework2021}, and we first noticed this problem and submitted our solution. As shown in Fig.\ref{fig:ROS1-framework}, we added the conversion for private and public IP addresses so that the robot and cloud server networks are under the same subnet. This step is necessary because the ROS is based on one ROS-master for communication. The original FogROS only provides a public IP connection, which is unfeasible in robotics application scenarios. This approach is also applied in the ROS2-based FogROS. The innovation of the Press-Elastic-Release node is to split one function node to generate a pressure node and a release node, and implement collaborative computing with the control of the elastic node. In fact, the code files of the press node and the released node are the same, but the execution content is dynamically changing with the control of the elastic node. Release nodes are deployed in the cloud, and press and elastic nodes are deployed in the robot. We illustrate the cloud hierarchy in Fig.\ref{fig:ROS1-framework}, taking AWS as an example as a cloud service provider. The robot generates image files and runs the nodes in EC2 of the VPC after local nodes are generated with launch files. ElasticROS does not destroy the node-level communication framework, and achieves algorithm-level collaborative computing by dynamically distributing the content of algorithm execution.
\subsection{Layer framework of ROS2-based ElasticROS}
\begin{figure}[H]
        \includegraphics[width=0.49\textwidth]{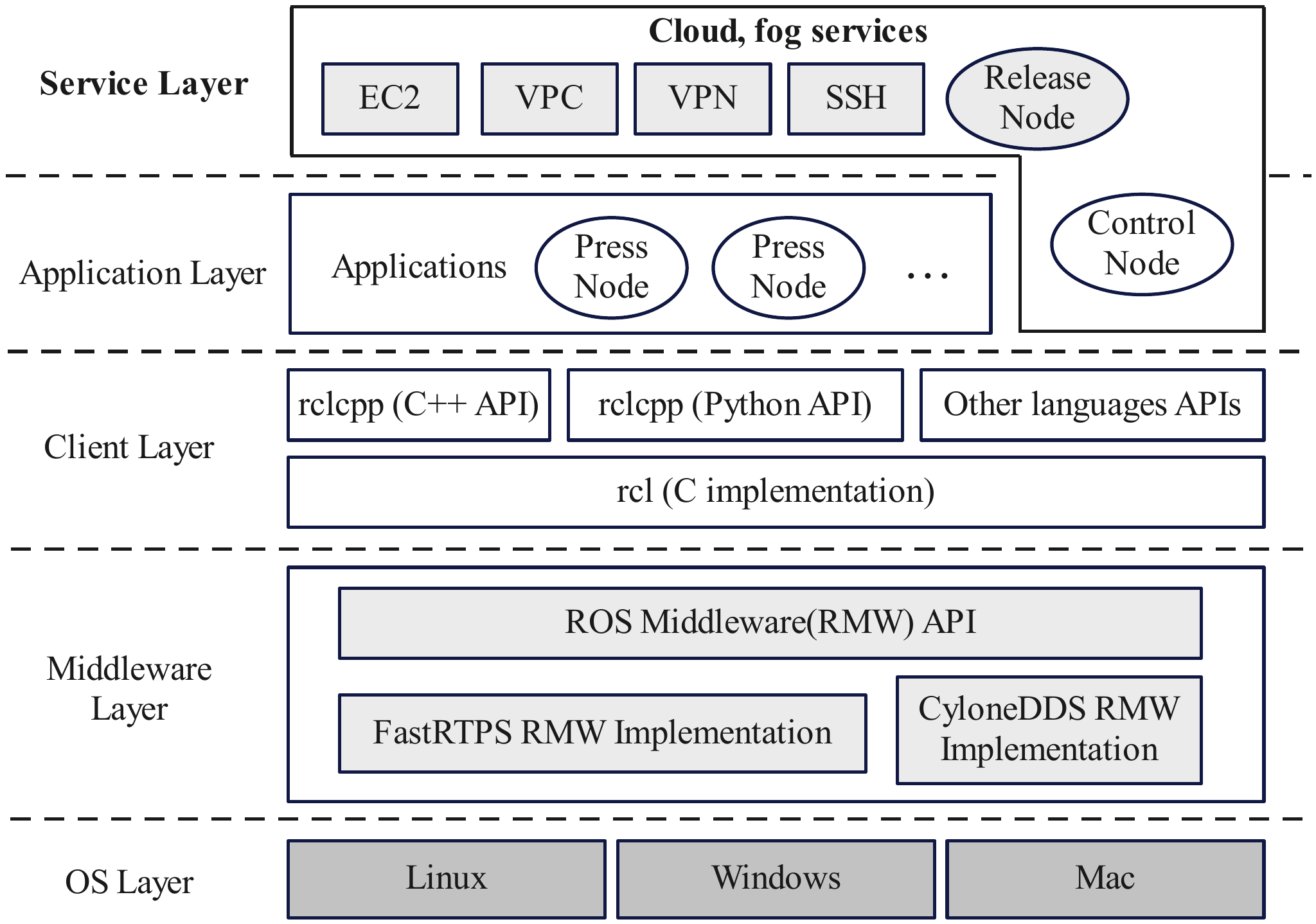}
        \caption{The layer framework of the ROS2-based ElasticROS}
        \label{ROS2}
\end{figure}
\begin{figure*}
    \centering
    \includegraphics[width=\textwidth]{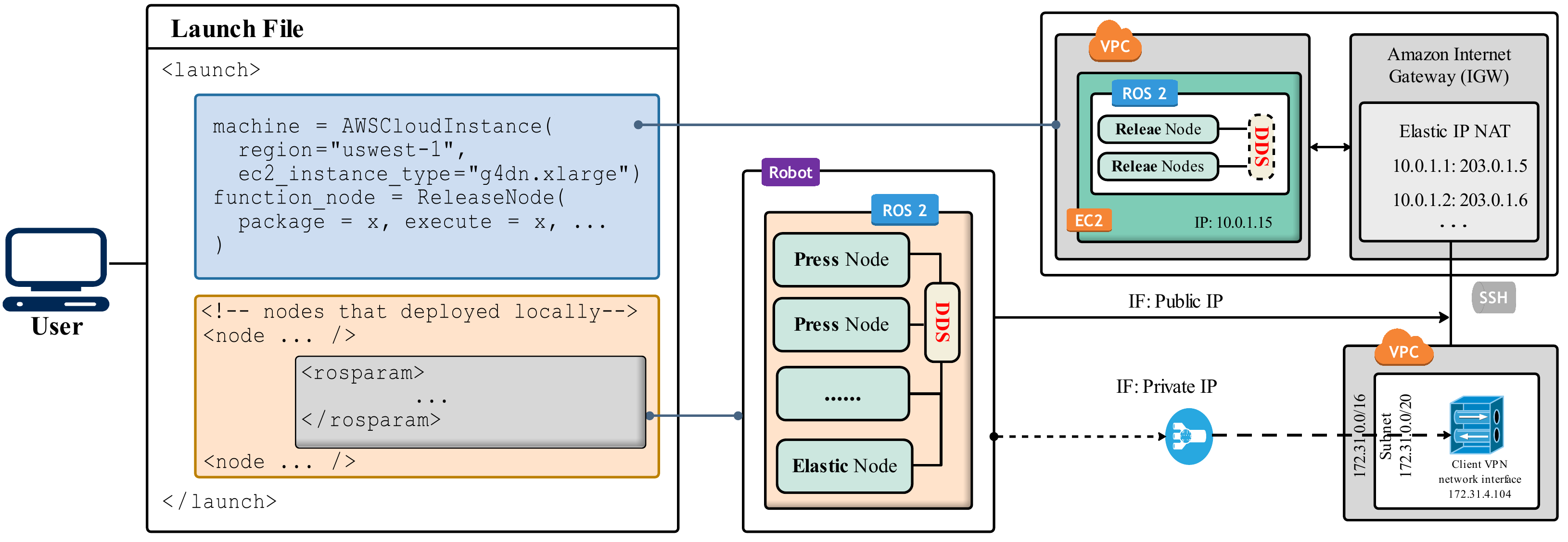}
    \caption{The network layout of ROS2-based ElasticROS.}
    \label{fig:ROS2net}
\end{figure*}
ROS is currently available in two versions, ROS and ROS2, so ElasticROS is also heterogeneous based on the two. The node-level communication framework FogROS also proposes FogROS2 for ROS2. ROS2-based ElasticROS inherits the advantages of ROS2 and FogROS2, and develops into algorithm-level collaborative computing. Data transfer between robot nodes in ROS-based ElasticROS are done through memory replication, and many system resources are wasted on self-communication. Real-time communication is not guaranteed. In addition, ROS-based ElasticROS managed communication between all nodes through a core master node, and a crash of the master node would cause the whole system to run incorrectly. ROS2-based ElasticROS is managed boot, so users can specify the order of node boot. In addition, ROS2-based ElasticROS also removes the ROS-master to improve the shortcomings of ROS's high dependence on the master node. In short, ROS2-based ElasticROS is more reliable, more sustainable, more resource efficient.

ROS-based ElasticROS is mainly built on the Linux system and mainly supports Ubuntu. As shown in Fig.\ref{ROS2}, ROS2-based ElasticROS adopts a new architecture and the underlying layer is based on the DDS communication mechanism, which supports embedded, distributed, and multi-operating systems.  The ROS2 based ElasticROS includes Linux, Windows, Mac, RTOS and even single-chip microcomputers that have no operating system. The core of ROS is the middle layer of anonymous publish-subscribe communication based on the nodes in the master. In contrast, ROS2-based ElasticROS uses DDS based on RTSP (Real-Time Publish-Subscribe) protocol as the middle layer. DDS (Data-Distribution Service) is an industry standard for publishing-subscription communications in real-time and embedded systems.  This point-to-point communication mode is similar to the middle layer of ROS1, but DDS does not need to communicate between two nodes through master nodes like ROS, which makes the system more fault-tolerant and flexible. As shown in Fig.\ref{ROS2}, the ROS2-based ElasticROS inherits the DDS module.

On top of the OS layer and Middleware Layer is the Client Layer, which provides APIs for the programming language. the Application Layer and Service Layer are the same as ElasticROS, adopting the Press-Elastic-Release Node model. ROS2 simplifies the process of configuring the network in terms of communication. All that is required is to ensure that the ROS\_DOMAIN\_ID is the same under the same LAN.
\subsection{Network layout of ROS2-based ElasticROS}
As present in Fig. \ref{fig:ROS2net}, the network deployment of ROS2-based ElasticROS is similar to ROS-based ElasticROS, and the Launch file is based on the FogROS2 implementation. The default middleware used by ElasticROS for communication here is DDS, since it is based on ROS2. In DDS, the main mechanism by which different logical networks share the physical network is called Domain ID. The default domain ID for all ROS2-based ElasticROS nodes is 0. To avoid interference between different groups of robots running ROS2-based ElasticROS on the same network, a different domain ID should be set for each group.

It is important to note that the optimization of the different ``Quality of Service options'' in the control transmission in DDS is different from the optimization of latency in this paper. For the case with latency as the goal, this paper is optimizing the amount of communication data as the optimization goal and does not consider the communication technology. In contrast, DDS is focused on optimizing the speed of transmission of data. Therefore, ElasticROS is capable of exhibiting better latency in experiments with addition of communication technologies of DDS. The understanding of ElasticROS and DDS improvements in real-time is sometimes mixed up when latency is the optimization goal. They are actually two different directions.
\subsection{Analysis of the message delivery process}
\begin{figure*}
    \centering
    \includegraphics[width=\textwidth]{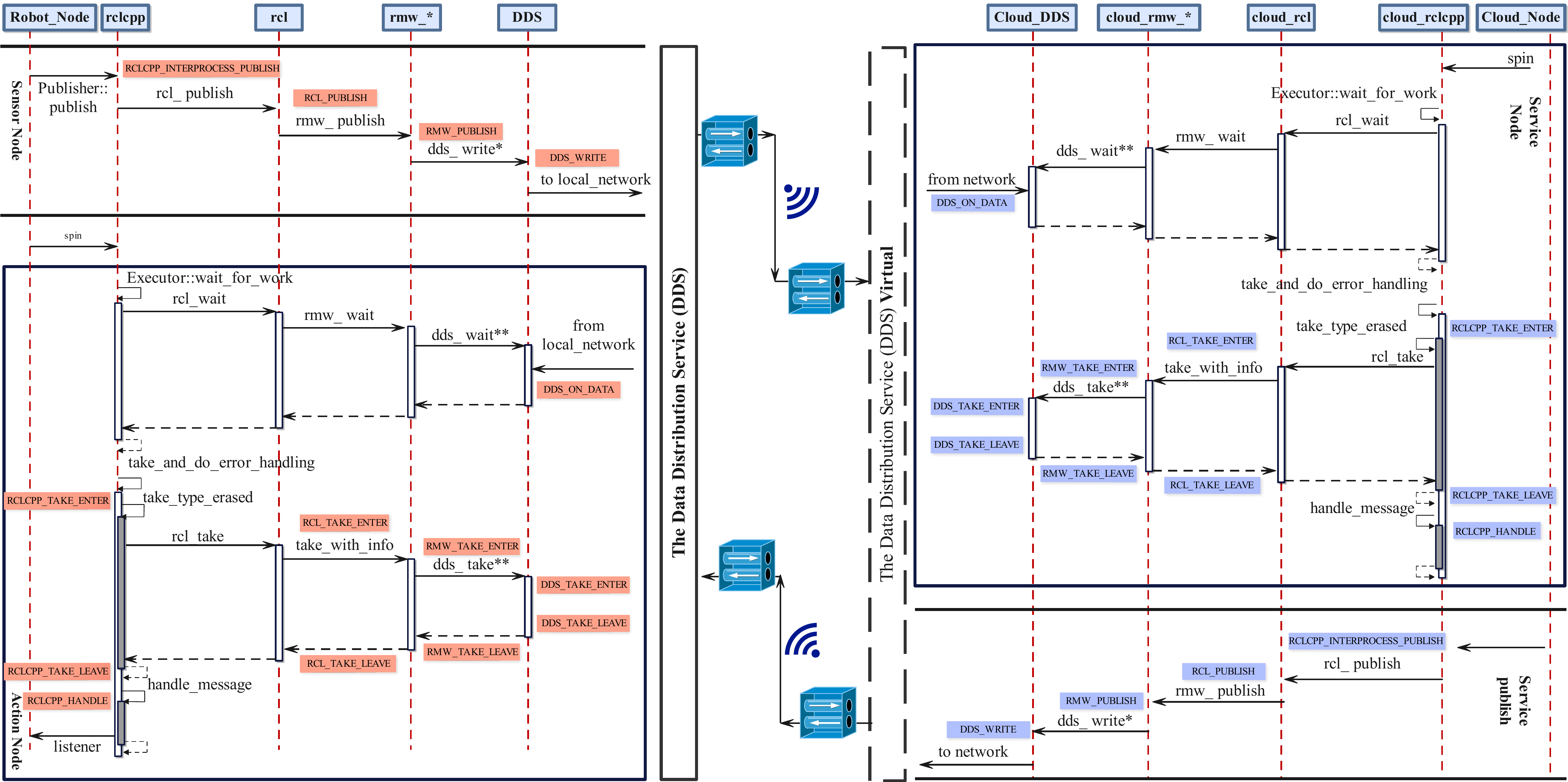}
    \caption{The messaging order diagram \cite{kronauer2021latency} in ElasticROS. ROS-based ElasticROS and ROS2-based ElasticROS are shown similarly here, except for the difference in DDS and ROS-master. The annotations point out the layers. The arrows indicate the direction of message delivery.}
    \label{fig:latency}
\end{figure*}
Fig.\ref{fig:latency} depicts the underlying function analysis of messages delivery in ElasticROS. The arrow annotations indicate the called function. The asterisk indicates a placeholder for the actual function name, as this depends on the middleware. Messages delivery relates to latency. For cloud robotics, latency is an issue that is discussed from time to time. Therefore, we further analyze latency here based on the flow analysis of messages delivery \cite{kronauer2021latency}.
\begin{itemize}
    \item DDS: This category contains only the latency required for the DDS to transmit over the internal network and for function calls to deliver messages.
    \item Subscriber rmw and Publisher rmw: Middleware that converts messages from ElasticROS to DDS messages. This is a necessary procedure to leverage the DDS functionality, but contains no delivery of the messages themselves. We attribute this category to the delay in the rmw layer.
    \item Publisher and Subscriber ROS2 Common: Overhead entailed by ROS2 that is independent of the middleware. 
    \item Rclcpp notification delay: The delay between the DDS notifying ElasticROS that new data is available and triggering its actual retrieval.
    \item Robot2Cloud and Cloud2Robot: Delay in message transmission between the robot and the server.
\end{itemize}

Of the overall ElasticROS latency, ROS2 is capable of us-level messaging latency internally, and ROS is capable of 10ms-level internal latency for data transfers of up to 5M. In contrast, the messaging latency of Robot2Cloud is much higher, accounting for more than 80\% of the overall latency. Therefore, optimizing the latency of ElasticROS begins with optimizing the Robot2Cloud module. We also conduct experiments with latency as an optimization objective in the experiment section in the paper. While cloud robotics tends to use latency as an optimization goal, the approach proposed in this paper is pervasive and capable of handling user-defined metrics. The key to achieve the ability of ElasticROS is the algorithm in the elastic node.
\section{Elastic collaborative computing algorithm for the Elastic Node}
\begin{figure}[!hpbt]
    \centering
    \includegraphics[width=0.49\textwidth]{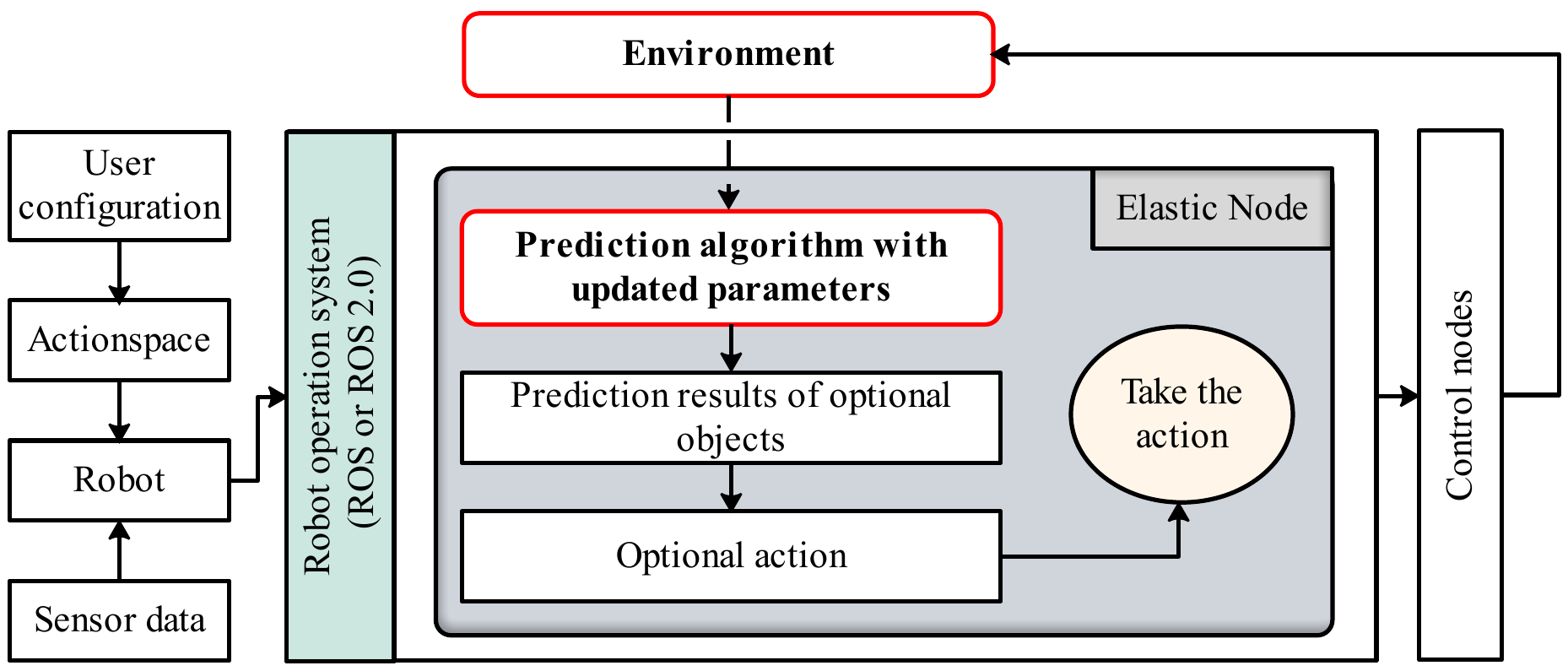}
    \caption{Process of computing of the elastic node. Arrows indicate the forwarding of data and boxes indicate computational modules. The robot operating system serves as an intermediary for the interaction between the robot and the environment. Feedbacks resulting from the execution of robot actions work on the elastic node function parameter updates. Computing process of the elastic node is cyclic and it is an online learning process.}
    \label{OnlineFramework}
\end{figure}
As shown in Fig. \ref{OnlineFramework}, the robot starts and generates the entire ElasticROS system after the user has configured the action space and evaluation metrics. The prediction function in the ElasticROS algorithm predicts the regret for each action in the action space. The regret is the gap between the result of the current action and the result of the optimal action. The elastic nodes select the optimal action to execute. The actual action outcomes are obtained after the actions are executed. The environment feeds the actual action results to the prediction function to update the parameters. This is in fact a form of online machine learning. In this approach, data becomes available sequentially. The data is then used at each step to update our best predictor for future results. It is different from batch learning techniques, which produce the best predictor by learning the entire training dataset at once. Online learning has appeared with many improvements since UCB was proposed in \cite{lai1985asymptotically}, including \cite{guo2019adalinucb, jamieson2014lil, zhang2021autodidactic, autwang2019q}.Our algorithm proof process is similar and based on UCB and these already proposed UCB-based algorithms. We propose a novel online learning algorithm applicable to ElasticROS for the actual working scenario of ElasticROS, and realize the elastic distribution of ElasticROS to obtain self-adaptive capability to environment changes.
\begin{figure}[!hpbt]
    \centering
    \includegraphics[width=0.49\textwidth]{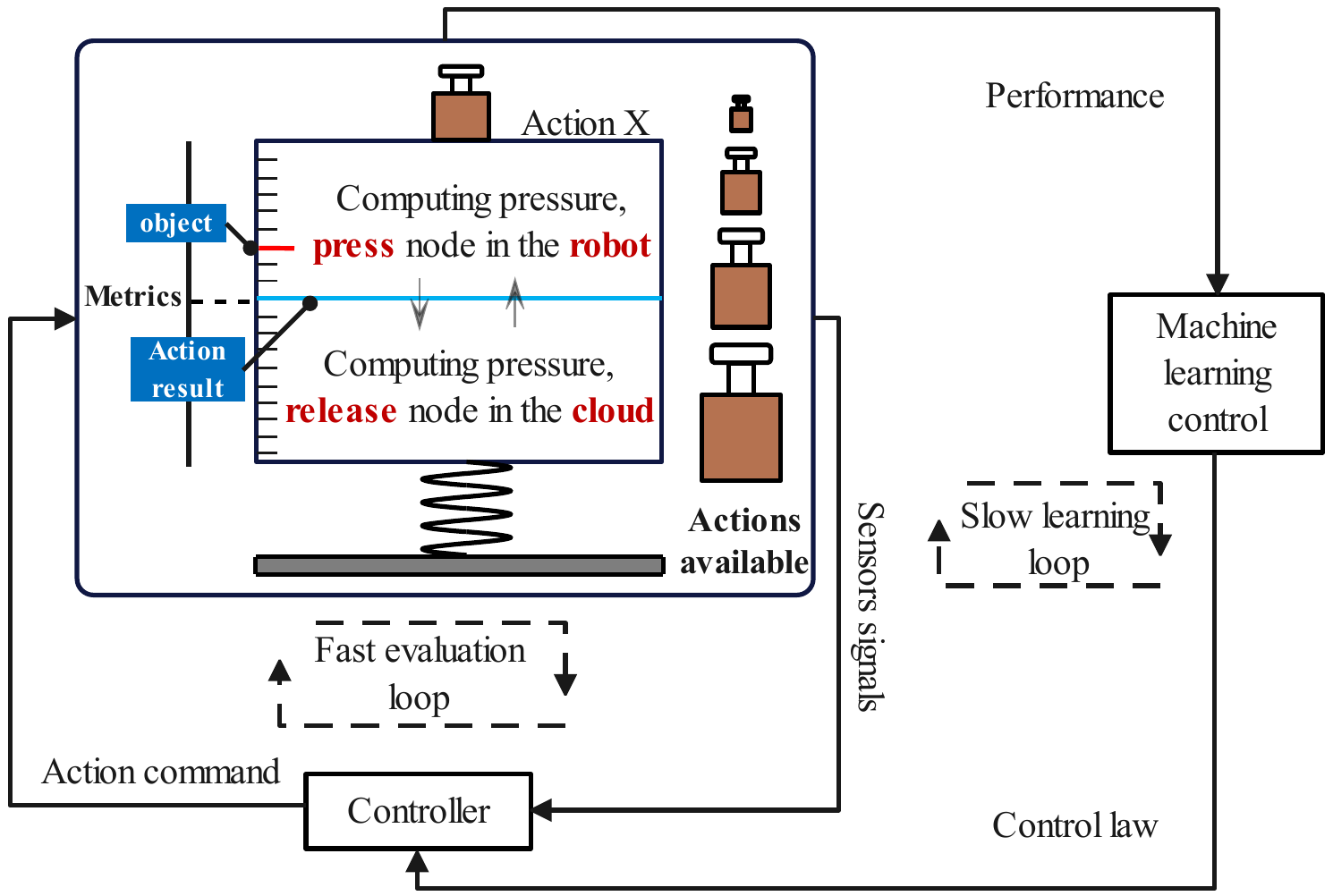}
    \caption{The analogous execution flow of the ElasticAction algorithm. The brown weights indicate the user-configured action space. The tick marks on the left side denote the results, where red indicates the optimization target and blue indicates the actual result. Press Node and Release Node are constant in the sum of their computations, but mutually exclusive.}
    \label{fig:elastic}
\end{figure}

Fig.\ref{fig:elastic} depicts the process of the proposed online learning based algorithm. We analogize the algorithm ElasticAction to a process of putting weights. Each weight is an action in the action space. The elastic node is to select an action such that the blue line meets the red bar of the optimal objective. The choice of action depends on the regret prediction of the algorithm for each action. The difference between the performance after executing the action and the performance predicted by the algorithm will effect the parameters update of the 
predictor. The proposed algorithm is inspired by traditional online learning. The basic idea of the ElasticAction algorithm is an online linear regression algorithm that incrementally updates the linear coefficients with continuous feedback. The  takes into account the information entropy of the predictions of the expected benefits of different actions when making decisions. The ElasticAction algorithm maintains two auxiliary variables $Q \in \mathrm{R}^{d \times d}$ and $\boldsymbol{p} \in \mathrm{R}^{d \times 1}$ for estimating the coefficients $\boldsymbol{\alpha}$. For each data frame $t, \alpha$ is estimated by $\hat{\boldsymbol{\alpha}}_{t}=Q_{t-1}^{-1} p_{t-1}$, and the action for frame $t$ is selected to be. In ElasticAction, The elastic point computing method is defined to be:
\begin{align}
\left\{
\begin{array}{lr}
    \vspace{1.5ex}
    \displaystyle
    a_{t}=\underset{a \in \mathcal{A}}{\arg \min }   \;E_{a}^{\mathrm{c}}+\hat{\boldsymbol{\alpha}}^{\top} \boldsymbol{v}_{a}-\beta \sqrt{\left(1-\sigma_{t}\right) \boldsymbol{v}_{a}^{\top} Q_{t-1}^{-1} v_{a}} \\\vspace{1.5ex}
    \displaystyle
    a_{t} \in A\left(E_{a}^{e} \longrightarrow E_{a}^{\text {real }}\right) \\\vspace{1.5ex}
    \displaystyle
    \sigma=u\left(\frac{-\sum_{i} \mathrm{P}\left(v_{i}\right) \log _{b} \mathrm{P}\left(v_{i}\right)}{Max}\right)
\end{array}
\right.
\end{align}
The $u$ in the third term in above formula demote a step function to determine the weight of key or none-key frames. The second term in the above formula represents a subset of the action space, depending on the relationship between the predicted and actual values, and updates the action space according to their relative relationship. For example, when the predicted delay is greater than the actual delay, we select the action in the action space where less data is transmitted. Also, this item is not force required. $\sigma_{t}$ denotes the information entropy, e.g. for image data, the action of ElasticAction is:
\begin{align}
\displaystyle
    a_{t}=& \underset{a \in \mathcal{Q}}{\arg \min }   \;E_{a}^{\mathrm{c}}+\hat{\boldsymbol{\alpha}}^{\top} \boldsymbol{v}_{a} \\ \nonumber
    & -\beta \sqrt{\left(1-\left({u\left(\frac{\left(-\sum_{i=0}^{255} {\frac{f(i, j)}{W \cdot H}} \log P_{i, j}\right)}{Max\sigma}\right)}\right)\right) \boldsymbol{v}_{a}^{\top} Q_{t-1}^{-1} v_{a}}
\end{align}
In the above functions to be minimized, the first term $E_{c}^{a}$ is the quantification of the on-device robot computing of the action $a$, which can be set to 0 or configured by the user. The second term $\hat{\boldsymbol{\alpha}}^{\top} \boldsymbol{v}_{t}$ is the predicted performance of the elastic action $a$ using the current estimate $\hat{\alpha}$. W and H denote the width and height of the image. $f(i,j)$ denotes the number of occurrences of the binary group $(i,j)$ in the whole image and $P_{i, j}$ denotes the pixel value.
 
In ElasticROS with elastic nodes, we leverage techniques that increase randomness to get rid of being trapped in a pure robot computing. It is crucial to obtain new information about the quantifiers configured by the user, so that $\alpha$ can be obtained. This idea of randomness is implemented in ElasticROS using forced sampling techniques. Specifically, a forced sampling sequence is defined for time steps where the total number of data from sensor nodes is $T$. 

Continuous acquisitions of new metric data related to the user configuration are key to enabling online learning, from which $\alpha$ can be updated. For example, for latency, nodes need to obtain the actual latency in order to update their predictions about the latency of data transmission. The idea of such randomness is implemented in ElasticROS using forced sampling techniques. Specifically, a forced sampling sequence is defined for a time step where the total number of data from the sensor nodes is $T$.
\begin{align}
\mathcal{S_F}=\left\{t \mid t=n T^{\frac{1}{{log_{I}}{T}}}, t \leq T, n=1,2, \ldots\right\},
\end{align}
where $I$ is a hyperparameter to determine the frequency. If the index $t$ of a sensor data belongs to $\mathcal{S_F}$, then ElasticROS sample an elastic point other than $a=P$. $P$ denotes the pure on-robot computing. Moreover, the force sampling frequency reduces gradually with increasing sampling interval in more common scenarios that $T$ are unknown.

The robot will start a fully local computing sub-process when it performs parameter updates. The execution module will execute based on the computed results of the local computation. These two modules are independent of each other. Elastic node parameter updates only occur when the environment changes and take up a small fraction of the entire robot execution cycle. So we are actually more concerned about the performance of the robot before and after the update, and we are more concerned about the results of the update process.

Algorithm \ref{Algorithm1} is the ElasticAction algorithm that is based on the above formulas.
\begin{algorithm}
\label{Algorithm1}
\caption{The ElasticAction Algorithm}
\SetKwInOut{Input}{input}\SetKwInOut{Output}{output}
	\Input{Construct context $v_{p}$ for candidate elastic points $\forall a \in \mathcal{A}$. Obtain robot computing metric estimate $E_{a}^{\mathrm{c}}, \forall a \in \mathcal{A}$. Determine forced sampling sequence $\mathcal{F}$. Initialize $Q_{0}=\gamma \  Y_{d}, p_{0}=0$.}
	\Output{Elastic point (action)}
	 \BlankLine
	 \For{each data frame $t=1,2, \cdots, T$}
	    {
	    Receive sensor data and assign weight $\sigma_{t}$
	    
        Compute current estimate $\hat{\alpha}_{t}=Q_{t-1}^{-1} p_{t-1}$.
        
        \For{each candidate elastic point $a \in \mathcal{A}$} 
	    {
	    Compute $\hat{E}_{a}^{\mathrm{e}}=\hat{\alpha}_{t}^{\top} v_{t}-\beta \sqrt{\left(1-\sigma_{t}\right) v_{p}^{\top} Q_{t-1}^{-1} v_{p}}$.
	    
	    \textbf{\Comment{Computing entropy}}
	    }
	    \If{$a_{t} \neq P$}
	    {
	    Observe $E_{a_{t}}^{\mathrm{e}}$ once the computing is done.
	    $Q_{t} \leftarrow Q_{t-1}+\boldsymbol{v}_{p_{t}} \boldsymbol{v}_{p_{t}}^{\top}, \boldsymbol{p}_{t} \leftarrow \boldsymbol{p}_{t-1}+\boldsymbol{v}_{p_{t}} E_{p_{t}}^{\mathrm{e}}.$
	    }
	    \Else{
	    $
	    \boldsymbol{Q}_{t}=\boldsymbol{Q}_{t-1}, \boldsymbol{p}_{t}=\boldsymbol{p}_{t-1}.
	    $
	    }
	    \If{$t \in \mathcal{F}$}
	    {
	    Choose $a_{t}=\arg \min _{a \in \mathcal{A}_{\{\neq P\}}} E_{a}^{c}+\hat{E}_{a}^{\mathrm{e}}$
	    }
	    \Else{
	        \If{Update}
	        {
	        $a_{t} \in A\left(E_{a}^{e} \longrightarrow E_{a}^{\text {real }}\right)$ 
	        
	        Then choose $a_{t}=\arg \min _{a \in \mathcal{A}} E_{a}^{\mathrm{c}}+\hat{E}_{a}^{\mathrm{e}}$
	        
	        \textbf{\Comment{Single direction update}}
	        }
	        \Else{
	        Choose $a_{t}=\arg \min _{a \in \mathcal{A}} E_{a}^{\mathrm{c}}+\hat{E}_{a}^{\mathrm{e}}$
	        }
	    }
	    }
 \end{algorithm}

Our algorithm is advanced in the following two ways.
\begin{itemize}
    \item The ElasticAction algorithm takes a single direction update when performing parameter updates. Compared with general online learning algorithms that update after randomly selecting in the action space, our algorithm makes full use of known information and is able to increase the update speed of the algorithm. ElasticROS makes the robot adapt to the new conditions as soon as possible.
    \item  The ElasticAction algorithm takes a step information-entropy function approach to parameters updating and action selection. The impacts of data frames are determined based on information entropies, and the computing is simplified using the step function.
\end{itemize}

The regret (i.e., the computing performance difference compared to an oracle algorithm that selects the optimal elastic point for all $T$ frames of ElasticAction, denoted by $R(T)$, satisfies: $\forall \epsilon \in(0,1)$, with probability at least $1-\epsilon$, $R(T)$ can be upper bounded by
\begin{align}
\displaystyle
    \max \left\{O\left(T^{0.5+\frac{1}{{log_{I}}{T}}} \log (T / \epsilon)\right), O\left(T^{1-\frac{1}{{log_{I}}{T}}}\right)\right\}
\end{align}
We then present the proof in the following.

The work analyzes the performance of ElasticAction by comparing it with the oracle solution, which knows precisely the ground-truth of the coefficient $\alpha^{*}$ and always chooses the best elastic point (action) $a_{t}^{*}$ to optimize the data computing for each $t$ frame. The performance is measured in terms of regret, the difference in the cumulative computing performance of all $T$ frames, which is described as follows:
\begin{equation}
    R=\sum_{t=1}^{T} E_{a_{t}}^{\mathrm{c}}+\alpha^{* \top} \boldsymbol{v}_{a_{t}}-E_{a_{t}^{*}}^{\mathrm{c}}-\alpha^{* \top} \boldsymbol{v}_{a_{t}^{*}}
\end{equation}
We first make some mild technical assumptions before obtaining the main results.
\begin{equation}
 \left\{
\begin{array}{lr}
    \vspace{1.5ex}
    0<E_{\text {non-key }}<E_{\text {key }}<1\\\vspace{1.5ex}
    \sigma_{t} \in\left\{\sigma_{\text {non-key }}, E_{\text {key }}\right\}\\\vspace{1.5ex}
    \forall a \in \mathcal{A},\left\|v_{a}\right\|_{2} \leq N_{v}\\\vspace{1.5ex}
    \left\|\alpha^{*}\right\|_{2} \leq C_{\alpha}&\\\vspace{1.5ex}
    \gamma \geq\left\{1, C_{\alpha}^{2}\right\}&
\end{array}
\right.
\end{equation}
The $x$ in formula (6) denotes the noise that satisfies the $\displaystyle N_{x}$-sub-Gaussian condition. To obtain the total $R$ clearly, we classify the sequence into three different types to analyze: Non-sampling data frames $\mathcal{S_N}$ : Normal frames that ElasticAction takes pure on-robot computing $a=\mathcal{P}$. Regular data frames $\mathcal{S_R}$ : Normal frames that ElasticAction takes an elastic point in $\mathcal{A}$. Forced data frames $\mathcal{S_F}$ : Forced sampling frames and ElasticAction takes an elastic point in $\displaystyle \mathcal{A}_{\{\neq P\}}$. Force samping data frames enables ElasticAction to observe $\displaystyle E_{a_{t}}^{\mathrm{e}}$ and update the $Q_{t}$ and $p_{t}$. These frames are interspersed during the operation of the ElasticAction. Let $\mathcal{T}_{F}=\left(t_{1}, \cdots, t_{F}\right)$ denote the subsequence of frames. Each $t_{f}$ is a sampling data frame. It is obvious that $F \leq T$. In the formulas, $Q_{f}, p_{f}$ are used to denote the matrix, the vector. The $\alpha_{f}$ denotes parameter estimation at the end of the $f$-th data frame.

\textbf{Lemma 1.} (The error bound of the predictor of ElasticAction.) With probability at least $1-\epsilon$, for any $\epsilon \in(0,1)$,  we can get: 

For all $a \in \mathcal{A}$ satisfies:
\begin{equation}
\left\{
\begin{array}{lr}
    \vspace{1.5ex}
    \left|\hat{\alpha}_{f}^{\top} v_{a}-\alpha^{* \top} v_{a}\right| \leq \beta \sqrt{\left(1-\sigma_{f}\right) v_{a}^{\top} Q_{m-1}^{-1} v_{a}} &\\
    \vspace{1.5ex}
    \displaystyle\sigma =\frac{-\sum_{i} \mathrm{P}\left(v_{i}\right) \log _{b} \mathrm{P}\left(v_{i}\right)}{Max}
    &\\
    \vspace{1.5ex}
    \displaystyle
    \beta=\frac{N_{\alpha}+N_{x} \sqrt{d \log \frac{1+M N_{x}^{2}}\displaystyle{\epsilon}}}{1-\displaystyle\left(u\left({\frac{-\sum_{i} \mathrm{P}\left(v_{i}\right) \log _{b} \mathrm{P}\left(v_{i}\right)}{Max}}\right)\right)}
\end{array}
\right.
\end{equation}

\begin{figure*}[!b]
    \rule[5pt]{\textwidth}{0.05em}
    \textbf{Proof.}
    \allowdisplaybreaks[3]
    \begin{align}
    \displaystyle
    \nonumber
    &\left|\hat{\alpha}_{f}^{\top} v_{a}-\alpha^{* \top} v_{a}\right|=\left|\left(\hat{\alpha}_{f}^{\top}-\alpha^{* \top}\right) v_{a}\right|=\left|\left(\hat{\alpha}_{f}^{\top}-\alpha^{* \top}\right) Q_{m-1}^{\frac{1}{2}} Q_{m-1}^{-\frac{1}{2}} v_{a}\right|=\left|\left(\hat{\alpha}_{f}^{\top}-\alpha^{* \top}\right) Q_{m-1}^{\frac{1}{2}} v_{a} Q_{m-1}^{-\frac{1}{2}}\right|\\[0.5\baselineskip]\nonumber &\leq\left\|\left(\hat{\alpha}_{f}^{\top}-\alpha^{* \top}\right) Q_{m-1}^{\frac{1}{2}}\right\|_{2}\left\|v_{a} Q_{m-1}^{-\frac{1}{2}}\right\|_{2}=\sqrt{\left(\hat{\alpha}_{f}-\alpha^{*}\right)^{\top} Q_{m-1}^{\frac{1}{2}} Q_{m-1}^{\frac{1}{2}}\left(\hat{\alpha}_{f}-\alpha^{*}\right)} \cdot \sqrt{v_{a}^{\top} Q_{t-1}^{-\frac{1}{2}} Q_{t-1}^{-\frac{1}{2}} v_{a}}\\[0.5\baselineskip]\nonumber
    &=\left\|\hat{\boldsymbol{\alpha}}_{f}-\alpha^{*}\right\|_{Q_{m-1}} \cdot \sqrt{v_{a}^{\top} Q_{m-1}^{-1} v_{a}}
    \leq\left(N_{\alpha}+N_{x} \sqrt{d \log \frac{1+M C_{x}^{2}}{\epsilon}}\right) \cdot \sqrt{v_{a}^{\top} Q_{m-1}^{-1} v_{a}}\\[0.5\baselineskip]\nonumber
    &=\left(\frac{N_{\alpha}+N_{x} \sqrt{d \log \frac{1+M C_{x}^{2}}{\epsilon}}}{1-\displaystyle\left({\frac{-\sum_{i} \mathrm{P}\left(v_{i}\right) \log _{b} \mathrm{P}\left(v_{i}\right)}{Max}}_{f}\right)}\right) \cdot \sqrt{\left(1-\sigma_{f}\right) v_{a}^{\top} Q_{m-1}^{-1} v_{a}}\\[0.5\baselineskip]
    & \le \left(\frac{N_\alpha + N_x \sqrt{d\log\frac{1 + M C_x^2}{\epsilon}}}{1 -\displaystyle \left(u\left({\frac{-\sum_{i} \mathrm{P}\left(v_{i}\right) \log _{b} \mathrm{P}\left(v_{i}\right)}{Max}}u\right)\right))}\right) \cdot \sqrt{(1 - E_m)v_a^\top Q^{-1}_{m-1}v_a}
    \end{align}
    \label{Formula11}
\end{figure*}

We conduct derivation in the formula (8), then we get the following:
\begin{align}
\displaystyle
\left|\hat{\alpha}_{f}^{\top} v_{a}-\alpha^{* \top} v_{a}\right| \leq \beta \sqrt{\left(1-\displaystyle\frac{-\sum_{i} \mathrm{P}\left(v_{i}\right) \log _{b} \mathrm{P}\left(v_{i}\right)}{Max}\right) v_{a}^{\top} Q_{m-1}^{-1} v_{a}}   
\end{align}

From the formulas we found that $Q_{t-1}$ is a symmetric positive infinite matrix, which is hold by the second equality. The first inequality holds by the Cauchy-Schwarz inequality. The second inequality holds by the Lemma 2 below. We then complete the proof based on the following:
\begin{align}
    \beta=\frac{N_{\alpha}+N_{x} \sqrt{d \log \frac{1+M N_{v}^{2}}{\epsilon}}}{1-\sigma(u=key)}
\end{align}

\textbf{Lemma 2.} For all $\epsilon \in(0,1)$, when $|x| \leq N_{x},\left\|\alpha^{*}\right\|_{2} \leq N_{\alpha},\left\|v_{a}\right\|_{2} \leq N_{v}$, with probability at least $1-\epsilon$, we have
\begin{align}
\left\|\hat{\boldsymbol{\alpha}}_{f}-\alpha^{*}\right\|_{Q_{m-1}} \leq N_{\alpha}+N_{x} \sqrt{d \log \frac{1+M N_{v}^{2}}{\epsilon}}
\end{align}
The $d$ in formulas is the dimension of the context.

It is proved to follow the fact that $\hat{\boldsymbol{\alpha}}_{t}$ is the result of ridge regression using the data samples collected in the sampling slot \cite{zhang2021autodidactic}. It is assumed that the noise is a sub-Gaussian condition. Theorem 2 in \cite{abbasi2011improved} gives a complete proof.

\textbf{Lemma 3.} (One-step regret of the robot action) $\forall m \geq 0$ let 
\begin{align}
    \displaystyle \beta=\frac{N_{\alpha}+N_{x} \sqrt{d \log \frac{1+M N_{v}^{2}}{\epsilon}}}{1-u\left(\frac{-\sum_{i} \mathrm{P}\left(v_{i}\right) \log _{b} \mathrm{P}\left(v_{i}\right)}{Max}\right)}
\end{align}
For the regret of one-step robot action, the following is satisfied:
\begin{equation}
\left\{
\begin{array}{lr}
    \vspace{1.5ex}
R_{t} \leq 2 \beta \sqrt{Q_{f}^{-1} v_{a}^{\top}  v_{a}}, \  When \ t \in \mathcal{S_R} &\\
    \vspace{1.5ex}
R_{t} \leq 3 \beta \sqrt{Q_{f}^{-1} v_{a}^{\top}  v_{a}}, \ When \ t \in \mathcal{S_R}
\end{array}
\right.
\end{equation}

The proof of the one-step regret of robot action is based on the fact that when ElasticAction takes pure on-robot computing with $a=P$, the regret of the computing is $E_{a}^{\mathrm{p}}$ that is calculated from user's configurations. We classify ElasticAction into four cases and analyzes the on-step regret.
\begin{enumerate}
    \item The optimal action is: $a^{*}=P$ while the elastic node takes action $a=P$. In this case, 
    \begin{align}
        \mathbb{E}\left(E_{a^{*}}^{f}-E_{a}^{c}\right)=0
    \end{align}
    Then, we get $R_{t}=0$.
    \item The ElasticAction algorithm takes action:\\
    $a \in A \{actions\ that\ with\ cloud\ computing\}$, while the optimal action is $a^{*} \in\{0,1, \cdots, n-1\}$. In this case, the one-step regret is present in the Formula (8) at the bottom of the page, 
    \begin{figure*}[b]
    \rule[5pt]{\textwidth}{0.05em}
    \begin{align}
    \nonumber
    &R_{t}=E_{a}^{c}+\alpha^{* \top} v_{a}-E_{p^{*}}^{f}-\alpha^{* \top} v_{p^{*}}
    \leq E_{a}^{c}+\alpha^{* \top} v_{a}-E_{p^{*}}^{f}-\hat{\alpha}_{f}^{\top} v_{p^{*}}^{*}+\beta \sqrt{\left(1-\displaystyle\left({\frac{-\sum_{i} \mathrm{P}\left(v_{i}\right) \log _{b} \mathrm{P}\left(v_{i}\right)}{Max}}_{f}\right)\right) v_{p^{*}}^{\top} Q_{m-1}^{-1} v_{p^{*}}}\\[0.5\baselineskip]\nonumber
    &=E_{a}^{c}+\alpha^{* \top} v_{a}-\left[E_{p^{*}}^{f}+\hat{\alpha}_{f}^{\top} v_{p^{*}}-\beta \sqrt{\left(1-\displaystyle\left({\frac{-\sum_{i} \mathrm{P}\left(v_{i}\right) \log _{b} \mathrm{P}\left(v_{i}\right)}{Max}}_{f}\right)\right) v_{p^{*}}^{\top} Q_{m-1}^{-1} v_{p^{*}}}\right] \\[0.5\baselineskip]\nonumber
    &\leq E_{a}^{c}+\alpha^{* \top} v_{a}-\left[E_{a}^{c}+\hat{\alpha}_{f}^{\top} v_{a}-\beta \sqrt{\left(1-\displaystyle\left({\frac{-\sum_{i} \mathrm{P}\left(v_{i}\right) \log _{b} \mathrm{P}\left(v_{i}\right)}{Max}}_{f}\right)\right) v_{a}^{\top} Q_{m-1}^{-1} v_{a}}\right] \\[0.5\baselineskip]\nonumber
    &=\alpha^{* \top} v_{a}-\hat{\alpha}_{f}^{\top} v_{a}+\beta \sqrt{\left(1-\displaystyle\left({\frac{-\sum_{i} \mathrm{P}\left(v_{i}\right) \log _{b} \mathrm{P}\left(v_{i}\right)}{Max}}_{f}\right)\right) v_{a}^{\top} Q_{m-1}^{-1} v_{a}} \\[0.5\baselineskip]
    &\leq 2 \beta \sqrt{\left(1-\displaystyle\left({\frac{-\sum_{i} \mathrm{P}\left(v_{i}\right) \log _{b} \mathrm{P}\left(v_{i}\right)}{Max}}_{f}\right)\right) v_{a}^{\top} Q_{m-1}^{-1} v_{a}}\\[0.5\baselineskip]\nonumber
    &\leq 2 \beta \sqrt{v_{a}^{\top} Q_{m-1}^{-1} v_{a}}
    \end{align}
    \end{figure*}
    The inequalities in the second and sixth lines of which hold according to formula (4). The inequality in the fourth row holds in our algorithm design.

\item 
    The optimal action is $a^{*}=P$ while the elastic node takes action $a \in\ A\ \{actions\ that\ with\ cloud\ computing\}$. The we get:
    \begin{align}
    \nonumber
    R_{t}=&\alpha^{*\top}v_{a}-E_{p}^{\mathrm{c}}+E_{a}^{\mathrm{c}} \\[0.5\baselineskip]
    &=\left(\alpha^{* \top} v_{a}-\hat{\alpha}_{f}^{\top} v_{a}\right)+\left(E_{a}^{\mathrm{c}}+\hat{\alpha}_{f}^{\top} v_{a}-E_{a}^{\mathrm{p}}\right) \\[0.5\baselineskip]\nonumber
    &\leq 2 \beta \sqrt{v_{a}^{\top} Q_{m-1}^{-1} v_{a}}
    \end{align}
    The inequality in formula (17) holds by Lemma 1. According to the discussion above, the one-step regret satisfies
    \begin{align}
    R_{t} \leq 3 \beta \sqrt{v_{a}^{\top} Q_{f}^{-1} v_{a}}
    \end{align}

\item 
      The optimal action is:\\
      $a^{*} \in A\{actions\ that\ with\ cloud\ computing\}$, while the elastic node takes action $a=P$. We firstly introduce an auxiliary action:\\
      $\hat{a} \in A\ {actions\ that\ with\ cloud\ computing}$. Therefore, we can get:
    \begin{align}
    \nonumber
    &R_{t}=E_{P}^{\mathrm{c}}-E_{a^{*}}^{\mathrm{c}}-\alpha^{* \top} v_{a^{*}}\\[0.5\baselineskip]\nonumber
    &=-E_{a^{*}}^{\mathrm{c}}-\alpha^{* \top} v_{a^{*}}+\hat{\alpha}_{f}^{\top} v_{\hat{a}}E_{P}^{\mathrm{c}}+E_{\hat{a}}^{\mathrm{a}}+\alpha^{* \top} v_{\hat{a}}\\[0.5\baselineskip]\nonumber
    &\qquad\ \ \ \ \ \ \ \ \ \ \ \ \ \ \ \ \ \ \ \ \ \ \ \ \ \ \ \ \ \ \ -\alpha^{* \top} v_{\hat{a}}-E_{\hat{a}}^{\mathrm{c}}-\hat{\alpha}_{f} v_{\hat{a}}\\[0.5\baselineskip]\nonumber
    &\leq E_{P}^{\mathrm{c}}-E_{\hat{a}}^{\mathrm{c}}-\hat{\alpha}_{f} v_{\hat{a}}+3 \beta \sqrt{v_{\hat{a}}^{\top} Q_{m-1}^{-1} v_{\hat{a}}}\\[0.5\baselineskip]
    &\leq 3 \beta \sqrt{Q_{m-1}^{-1} v_{\hat{a}}^{\top} v_{\hat{a}}}
    \end{align}
    The inequality in the third line holds by the Lemma 1 and Case 2) in Formula (19) in the following in this case. The last inequality holds because of the Formula (22) in this case.
    \begin{align}
        E_{P}^{\mathrm{c}}-E_{\hat{a}}^{\mathrm{c}}-\hat{\alpha}_{f} v_{\hat{a}} \leq 0
    \end{align}
\end{enumerate}

\textbf{Lemma 4.} Assume $\left\|v_{a}\right\|_{2} \leq N_{x}$ and the minimum eigenvalue of $Q_{0}$ satisfies:\\
$\lambda_{\min }\left(Q_{0}\right) \geq \max \left\{1, C_{x}^{2}\right\}$. Then, we have
\begin{align}
\nonumber
&\sum_{t=1}^{T} v_{a}^{\top} Q_{t-1}^{-1} v_{a} \leq 2 \log\left(\frac{\operatorname{det}\left(Q_{F}\right)}{\operatorname{det} I_{d}}\right)\\[0.5\baselineskip]
&\leq 2 d\left[\log \left(\gamma+\frac{M C_{x}^{2}}{d}\right)-\log \gamma\right]
\end{align}
The complete proof follows Lemma 11 of \cite{abbasi2011improved}.

We then get the total regret of the the regular sampling sequence $\mathcal{S_R}$ with Lemma 3 and Lemma 4.
\begin{align}
\nonumber
&R_{\mathcal{S_R}}=\sum_{t=1}^{T} R_{t} \mathbf{1}\{t \in \mathcal{S_R}\} \leq \sqrt{F \sum_{f=1}^{F} R_{t_{f}}^{2} \mathbf{1}\{t \in \mathcal{S_R}\}}\\[0.5\baselineskip]\nonumber
&\leq \sqrt{4 F \beta^{2} \sum_{f=1}^{F} v_{f, a}^{\top} Q_{f}^{-1} v_{f, a}}\\[0.5\baselineskip]\nonumber
&\leq 2 \beta \sqrt{2 F d\left[\log \left(\gamma+\frac{F N_{v}^{2}}{d}\right)-\log \gamma\right]}\\[0.5\baselineskip]
&\leq 2 \beta \sqrt{2 T d\left[\log \left(\gamma+\frac{T N_{v}^{2}}{d}\right)-\log \gamma\right]}=2 G(T)
\end{align}

We obtain the third inequality from Lemma 4 and the fourth inequality holds according to the fact $M\leq T$. The first inequality holds according to Jensen's inequality, and the second inequality holds according to Lemma 3 and relaxing the indicator function $\mathbf{1}\{t \in \mathcal{S_R}\}$.

We then analyze the total regret incurred in non-sampling sequences $\mathcal{S_N}$:
\begin{align}
\nonumber
    R_{\mathcal{S_N}}&=\sum_{t=1}^{T} R_{t} \mathbf{1}\{t \in \mathcal{S_N}\}\\[0.5\baselineskip]\nonumber
    &\leq T^{\frac{1}{{log_{I}}{T}}} \sum_{f=1}^{F} R_{f}
    \leq T^{\frac{1}{{log_{I}}{T}}} 3 \beta \sqrt{v_{a}^{\top} Q_{t}^{-1} v_{a}}\\[0.5\baselineskip]
    &=3 T^{\frac{1}{{log_{I}}{T}}} \cdot G(T)
\end{align}
Then, we analyze the total regret of the forced sampling sequence $\mathcal{S_F}$.

\begin{equation}
    R_{\mathcal{S_F}}=\sum_{t=1}^{T} R_{t} \mathbf{1}\{t \in \mathcal{S_F}\} \leq T^{1-\frac{1}{{log_{I}}{T}}_{\Delta_{\max }}}
\end{equation}
In formula (22) $\Delta_{\max }$ denotes the maximum metrics gap between robot computing and other elastic points. we obtain the following by combining these regret bounds.
\begin{align}
\nonumber
R_{\text {total }}=R_{\mathcal{S_R}}&+R_{\mathcal{S_N}}+R_{\mathcal{S_F}}\\[0.5\baselineskip]
&\leq\left(2+3 T^{\frac{1}{{log_{I}}{T}}}\right) G(T)+T^{1-\frac{1}{{log_{I}}{T}}} \Delta_{\max }
\end{align}
\begin{align}
\nonumber
    G(T)&=\frac{N_{\alpha}+N_{x} \sqrt{d \log \frac{1+T N_{v}^{2}}{\epsilon}}}{1-\sigma(u=key)}\qquad\qquad\qquad\qquad\qquad\qquad\\[0.5\baselineskip]
    &\qquad\qquad\qquad\cdot \sqrt{2 T d\left[\log \left(\gamma+\frac{T N_{v}^{2}}{d}\right)-\log \gamma\right]}
\end{align}

In the above formula, $\displaystyle G(T)=O\left(T^{0.5} \log (T / \epsilon)\right)$. Thus, Theorem 1 shows that the regret bound of the algorithm in the elastic node is sublinear in $T$, or $\max \left\{O\left(T^{0.5+\frac{1}{{log_{I}}{T}}} \log (T / \epsilon)\right), O\left(T^{1-\frac{1}{{log_{I}}{T}}}\right)\right\}$ by choosing $\frac{1}{{log_{I}}{T}} \in(0,0.5)$, $0<I<T^{0.5}$.

For the more common case where T is unknown, we can set T to the time interval of the last term and sum to obtain:
\begin{align}
\nonumber
    &G(T_{total})=O\left(T^{0.5} \log (T / \epsilon)\right)+O\left(\frac{T}{r}^{0.5} \log (\frac{T}{r} / \epsilon)\right)+\\[0.5\baselineskip]
    &...+O\left(\frac{T}{r^{n-1}}^{0.5} \log (\frac{T}{r^{n-1}} / \epsilon)\right)
\end{align}

\begin{figure*}[b]
\rule[5pt]{\textwidth}{0.05em}
\begin{align}
\nonumber
&R_{\text {total }}\leq
\left(2+3(\frac{T}{r^{n-1}})^{\frac{1}{{log_{I}}{T}}}\right)\frac{N_{\alpha}+N_{x} \sqrt{d \log \frac{1+(\frac{T}{r^{n-1}}) N_{x}^{2}}{\epsilon}}}{1-\sigma(u=key)} \cdot \sqrt{2 T d\left[\log \left(\gamma+\frac{(\frac{T}{r^{n-1}}) N_{x}^{2}}{d}\right)-\log \gamma\right]}+(\frac{T}{r^{n-1}})^{1-\frac{1}{{log_{I}}{T}}} \Delta_{\max }\\[0.5\baselineskip]\nonumber
&+\left(2+3(\frac{T}{r^{n-2}})^{\frac{1}{{log_{I}}{T}}}\right)\frac{N_{\alpha}+N_{x} \sqrt{d \log \frac{1+(\frac{T}{r^{n-2}}) N_{x}^{2}}{\epsilon}}}{1-\sigma(u=key)} \cdot \sqrt{2 T d\left[\log \left(\gamma+\frac{(\frac{T}{r^{n-2}}) N_{x}^{2}}{d}\right)-\log \gamma\right]}+(\frac{T}{r^{n-2}})^{1-\frac{1}{{log_{I}}{T}}} \Delta_{\max }\\[0.5\baselineskip]\nonumber
&+......+\\[0.5\baselineskip]\nonumber
&+\left(2+3(\frac{T}{r})^{\frac{1}{{log_{I}}{T}}}\right)\frac{N_{\alpha}+N_{x} \sqrt{d \log \frac{1+(\frac{T}{r}) N_{x}^{2}}{\epsilon}}}{1-\sigma(u=key)}\cdot \sqrt{2 T d\left[\log \left(\gamma+\frac{(\frac{T}{r}) N_{x}^{2}}{d}\right)-\log \gamma\right]}+(\frac{T}{r})^{1-\frac{1}{{log_{I}}{T}}} \Delta_{\max }\\[0.5\baselineskip]
&+\left(2+3T^{\frac{1}{{log_{I}}{T}}}\right)\frac{N_{\alpha}+N_{x} \sqrt{d \log \frac{1+T) N_{x}^{2}}{\epsilon}}}{1-\sigma(u=key)}
\cdot \sqrt{2 T d\left[\log \left(\gamma+\frac{T) N_{x}^{2}}{d}\right)-\log \gamma\right]}+(T)^{1-\frac{1}{{log_{I}}{T}}} \Delta_{\max }
\end{align}
\end{figure*}

The extended reasoning is shown in Formula (27) at the bottom of the page. We can naturally obtain that each term and the sum is also sublinear. The conclusion is the same as in the case where T is known.

According to the above proof and theorems, by taking $\frac{1}{{log_{I}}{T}} \in(0,0.5)$, $0<I<T^{0.5}$. the regret bound is sublinear in $T$, implying that the average computing metrics achieves the best possible performance when $T \rightarrow \infty$. For a finite $T$, this bound also gives a characterization of the convergence speed of $\frac{1}{{log_{I}}{T}}$. In addition, by take $\frac{1}{{log_{I}}{T}}=0.25$, the order of the regret bound is minimized at:
\begin{align}
  O\left(T^{0.75} \log (T)\right)
\end{align}

Here, we have completed the introduction and proof of the algorithm. The convergence of the algorithm guarantees the robustness of ElasticROS for elastically cooperative computing. Next, we experimentally validate ElasticROS. ElasticAction, similar to general Online Learning approaches, manages the balance between exploration and utilization through hyperparameters. Since we adopt a single-way update strategy, the convergence speed is steadily increased.
\section{Experiments}
In this section, we design experiments to fully validate ElasticROS. We configure latency as the the optimal objective, as it is the major concern in cloud robotics. The three main questions we want to answer are as follows: \textbf{1)} Does ElasticROS improve the latency performance for cloud robotics? \textbf{2)} Does ElasticROS improves robots’ performances for other computing metrics (e.g., CPU usage, power consumption) in common robotic tasks? \textbf{3)} Does ElasticROS is capable of being elastic and adaptive in adjusting its computing strategies in dynamic conditions? In order to provide justified answers to these questions, we compare pure robot computing, node-level computing and ElasticROS in three different robotic tasks with varying limitations of bandwidth to show the generality.

The three robotic tasks are SLAM, grasping, and human-robot dialogue. ElasticROS performs an elastic collaborative computing upgrade for one of the computing nodes in the task and quantifies the performance. In the SLAM task, we calculate computing distribution strategies of ElasticROS with different bandwidths to demonstrate its elasticity. In the grasping task, we supplemented an online bandwidth glitch challenge to demonstrate the adaptive capability of ElasticROS. In the human-robot dialogue task, we added a CPU Usage glitch challenge. In summary, all these are to verify that ElasticROS is capable of adapting no matter what conditions it faces.
\subsection{Experimental setup}
Our experiments are deployed on a small robot with a Jetson Nano (Robot) and a server (Cloud) that is a computer configured with an RTX 3090 graphics card. The robot and the server are connected through a wireless network. We selected a single DNN node in each task system for computing improvement with ElasticROS in our experiments. The advantages of selecting DNN nodes include their commonness, ease of configuring the action space, and understanding the amount of data transfer.

Latency is a primary concern for cloud robotics, so our experiments are configured to have latency as an optimization objective. We use Wondershaper \cite{hubert2021wondershaper} to change the bandwidth, which is one of the easiest and fastest ways to limit the bandwidth of a Linux system's Internet or local network. In our experiments, the data transfer latency from the robot to the server is the main factor affecting latency, while the reverse transfer has little effect, so the main statistic is the former. Experiments of node-level system of FogROS and pure robot computing were performed under the same hardware and software conditions.

Notation of experimental section: 1) In practice, the algorithm-level ElasticROS is not limited by the DNN functions. It can apply to all the computing nodes. And ElasticROS is also not limited by number of nodes, but depends on the user-configured action space. 2) In the experimental statistics below, the CPU and power consumption for ElasticROS during the parameter update period (i.e., the ``O'' period) are for the actual execution of the on-robot computing module, and the subprograms of the exploration module are not included in the statistics. Here only the on-robot computing is used as a criterion. It is reasonable because this phase occupies only a small percentage and we are more concerned with the performance when the parameters are updated.
\subsection{Experiments: SLAM}
\begin{figure}
    \centering
    \includegraphics[width=0.49\textwidth]{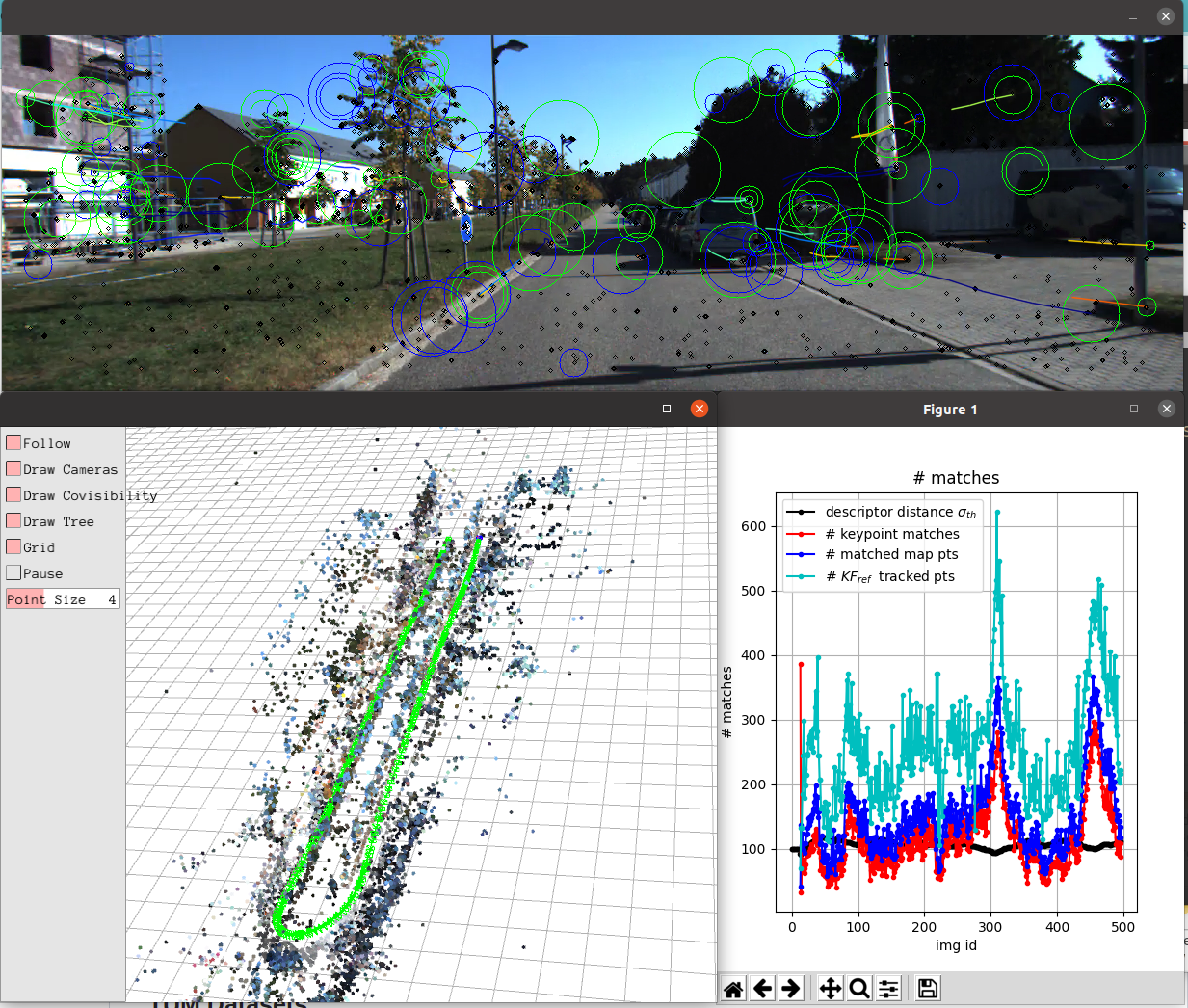}
    \caption{The experimental scenario of SLAM. Simultaneous localization modules are computing via ElasticROS.}
    \label{fig:slam}
\end{figure}
\begin{figure}
    \centering
    \includegraphics[width=0.49\textwidth]{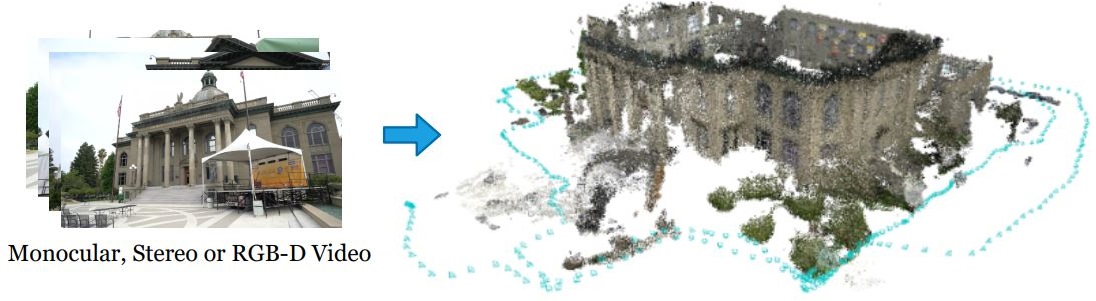}
    \caption{The End-to-end SLAM approach present in \cite{teed2021droid}, ElasticROS potential application scenarios for high computing power required SLAM tasks.}
    \label{fig:potentalSLAM}
\end{figure}
\begin{figure*}
\centering
     \subfloat{\includegraphics[width=0.84\textwidth]{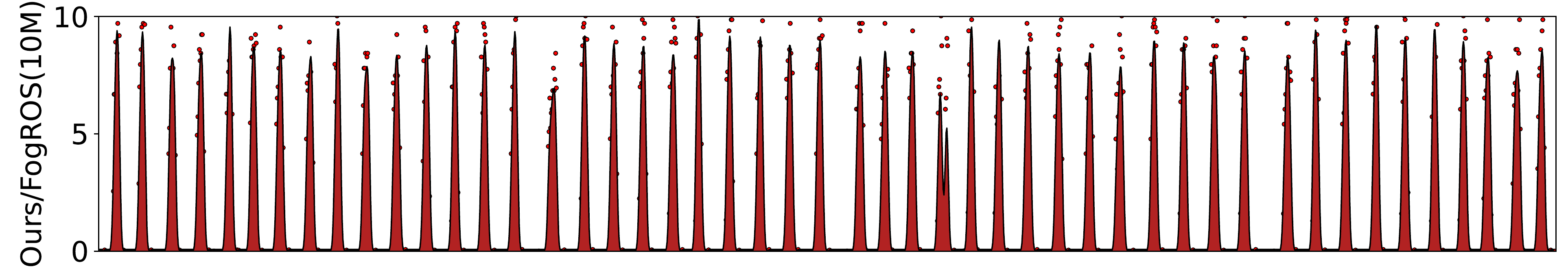}}
     	\hfill
     \subfloat{\includegraphics[width=0.84\textwidth]{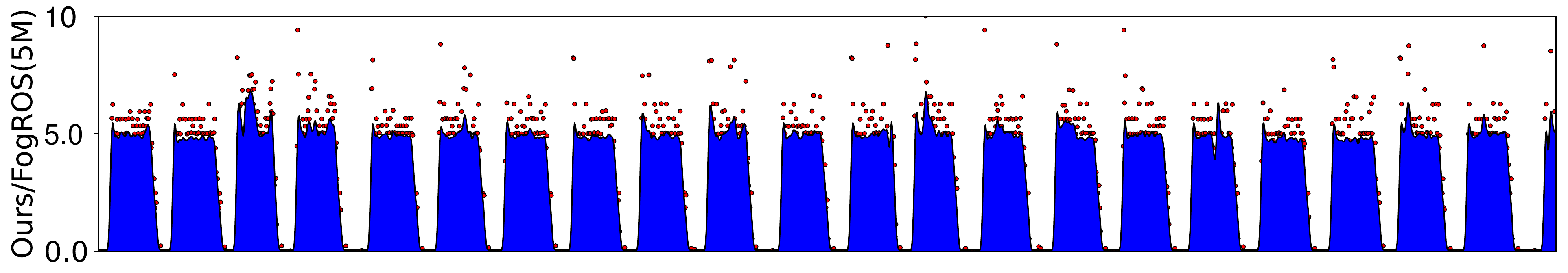}}
     	\hfill
     \subfloat{\includegraphics[width=0.84\textwidth]{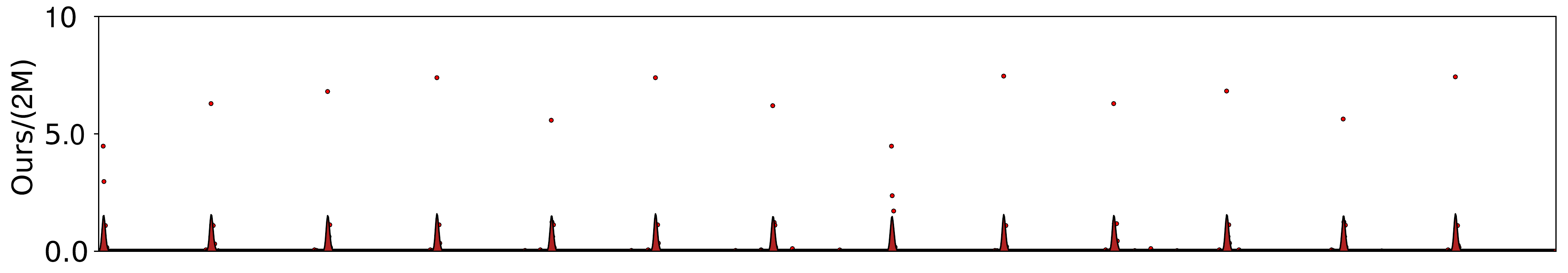}}
     	\hfill
     \subfloat{\includegraphics[width=0.84\textwidth]{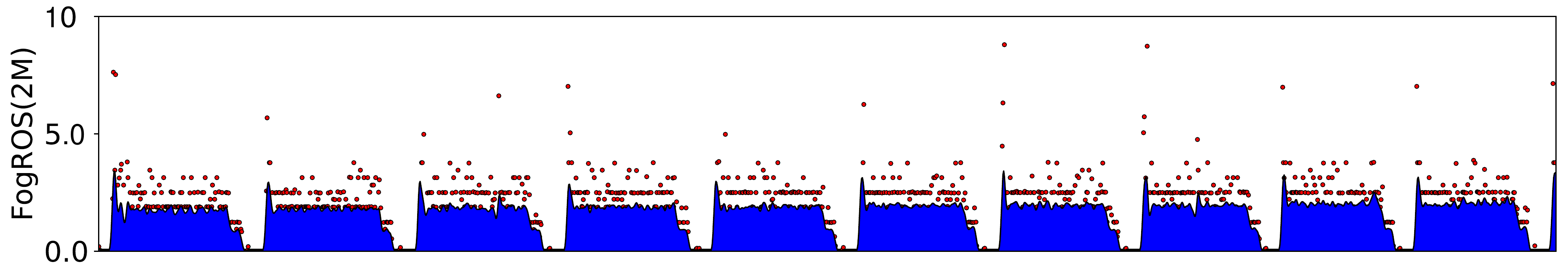}}
     	\hfill
     \subfloat{\includegraphics[width=0.84\textwidth]{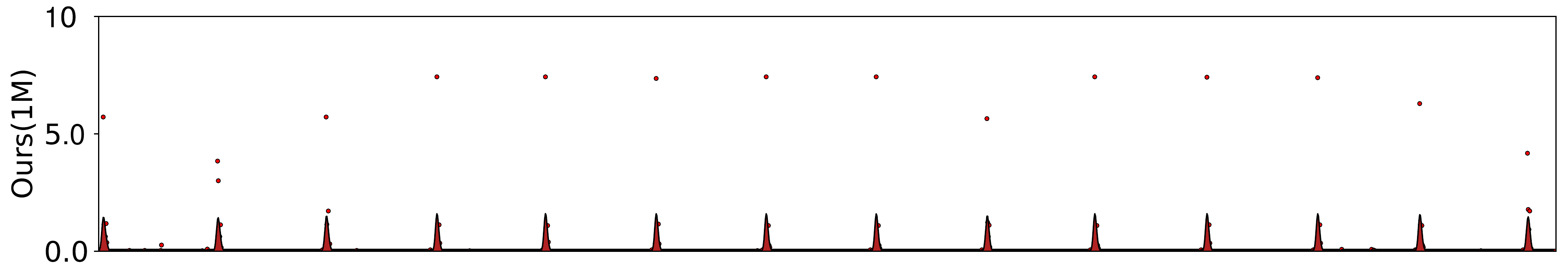}}
     	\hfill
     \subfloat{\includegraphics[width=0.84\textwidth]{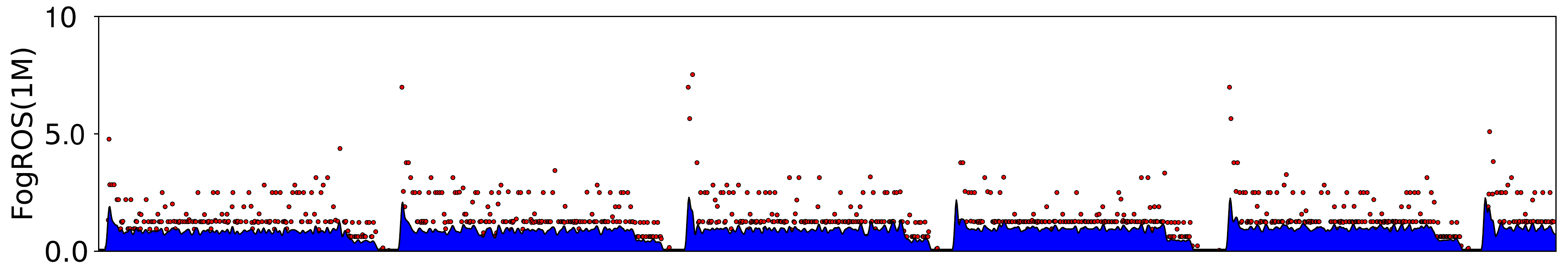}}
     	\hfill
     \qquad\subfloat{\includegraphics[width=0.84\textwidth]{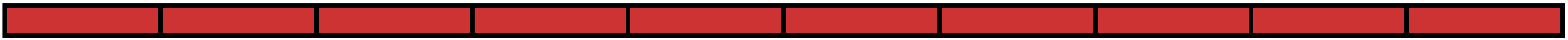}}
     \caption{Comparison of the results of FogROS and ElasticROS. The horizontal axis is the time axis and the vertical axis is the network speed. The color overlay part is fitted to the scatter, and the scale of the vertical axis is after simultaneous adjustment according to the fit. The color overlay also reflects the data transmission volume. The data transmission frequency reflects the system performance. The bottom red line is a fully robot computing rate.}
    \label{fig:SLAM-figure-compare}
\end{figure*}
SLAM is one of the common tasks for robots. In this experiment, we perform simultaneous localization experiments in SLAM leveraging pySLAM \cite{pyslam} as present in Fig.\ref{fig:slam}. We fuse end-to-end Homograph estimation into pySLAM and optimize the module leveraging ElasticROS. Fig. \ref{fig:potentalSLAM} shows the experimental scenario. Meanwhile, we provide Fig. \ref{fig:potentalSLAM}, an end-to-end SLAM work, which is a potential application scenario. Because end-to-end SLAM approaches have more accurate performance than traditional SLAM approaches, but high computational resource requirements hinder their applications. Therefore, we also provide the figure to demonstrate the practical value of ElasticROS.

Fig. \ref{fig:SLAM-figure-compare} depicts the performance comparison between ElasticROS and FogROS in the SLAM task, and also the performance comparison between the algorithm-level system and the node-level system. The first row of the figure shows the computation with a network speed of 10M/s. In this case of sufficient resources for the network speed, ElasticROS and FogROS take the same computing strategy, i.e., full cloud computing. A larger performance improvement is achieved compared to the fully local computation in the last row. This also reflects the feasibility of cloud robotics. The second row shows the computing in the case of network speed 5M/s, where the network resources are limited, ElasticROS also takes a full cloud computing. The third and fourth rows present a comparison for the case of network speed at 2M/s. ElasticROS takes different computing strategies compared to FogROS, where ElasticROS chooses actions that transfer less data. It has allocated the computation elastically according to the resources, resulting to a better performance. The fifth and sixth rows show a comparison with a network speed of 1M/s, a scenario with highly constrained network speed, where ElasticROS embodies a more superior performance improvement, more than twice the performance of the node-level FogROS.

\linespread{1.5}
\begin{table*}[htbp]
  \centering
  \caption{Comparison of experimental results for pure robot-computing, FogROS and ElasticROS in the SLAM task.}
    \resizebox{\textwidth}{!}{
    \begin{tabular}{cccccccccc}
    \toprule
    \multirow{2}[4]{*}{\makecell[c]{Metrics/\\ Speed}} & \multicolumn{3}{c}{Latency(s)} & \multicolumn{3}{c}{CPU  (relative usage)} & \multicolumn{3}{c}{Power consumption (mw/robot frame)} \\
\cmidrule(lr){2-4} \cmidrule(lr){5-7} \cmidrule(lr){8-10} & FogROS & \textbf{ElasticROS} & Robot & FogROS & \textbf{ElasticROS} & Robot & FogROS & \textbf{ElasticROS} & Robot \\
    \midrule
    512K-up limitation & 10.18   & {\color{green}{$\downarrow$}}\quad \textbf{1.99}\quad {\color{green}{$\downarrow$}}   & 2.61   & 1+17\%   & {\color{green}{$\downarrow$}}\quad \textbf{62\%} \quad{\color{green}{$\downarrow$}}  & 96\%   & 9614 &{\color{green}{$\downarrow$}}\quad\textbf{3607}\quad{\color{green}{$\downarrow$}}   & 4617 \\
    1M-up limitation& 5.20    & {\color{green}{$\downarrow$}}\quad \textbf{1.92}\quad {\color{green}{$\downarrow$}}   & 2.61   & 61\%   & {\color{green}{$\downarrow$}}\quad \textbf{58\%} \quad{\color{green}{$\downarrow$}}  & 96\%   & 4911 &{\color{green}{$\downarrow$}}\quad\textbf{3481}\quad{\color{green}{$\downarrow$}}   & 4610 \\
    2M-up limitation& 2.72    & {\color{green}{$\downarrow$}}\quad \textbf{1.81}\quad {\color{green}{$\downarrow$}}   & 2.62   & 32\%   &{\color{red}{$\uparrow$}}\quad \textbf{55\%} \quad {\color{green}{$\downarrow$}}   & 95\%   & 2559 &{\color{red}{$\uparrow$}}\quad\textbf{3269}\quad{\color{green}{$\downarrow$}}   & 4621 \\
    5M-up limitation& 0.97   & {\color{blue}{$=$}}\quad \textbf{0.97}\quad {\color{green}{$\downarrow$}}   & 2.62   & 11\%   &{\color{blue}{$=$}}\quad\textbf{11}\%\quad {\color{green}{$\downarrow$}}   & 96\%   & 912   &{\color{blue}{$\approx$}}\quad\textbf{911}\quad {\color{green}{$\downarrow$}} & 4601 \\
    20M-up limitation& 0.27   & {\color{blue}{$=$}}\quad \textbf{0.27}\quad {\color{green}{$\downarrow$}}   & 2.61   & 3.1\%   &{\color{blue}{$=$}}\quad\textbf{3.1}\%\quad{\color{green}{$\downarrow$}}& 97\%   & 255   & {\color{blue}{$\approx$}}\quad\textbf{271}\quad {\color{green}{$\downarrow$}}& 4611 \\
    1M-down limitation& 0.30   & {\color{blue}{$=$}}\quad\textbf{0.30}\quad{\color{green}{$\downarrow$}}   & 2.61   & 3.3\%   &{\color{blue}{$=$}}\quad\textbf{3.3\%}\quad{\color{green}{$\downarrow$}}& 96\%   & 282   & {\color{blue}{$\approx$}}\quad\textbf{297}\quad {\color{green}{$\downarrow$}}& 4611 \\
    \bottomrule
    \end{tabular}%
    }
  \label{tab:slam-table}%
\end{table*}%

Table \ref{tab:slam-table} shows a comparison of experimental data results for pure robot computing, FogROS, and ElasticROS. Also, we compare CPU usage and power consumption, noting that these two metrics are relative calculations that take computation frequency into account. The bolded data in the table shows the performance of ElasticROS, and the arrows on its left side indicate the comparison with FogROS and the arrows on the right side indicate the comparison with pure robot computing. Red indicates an increase in the result of the metric where it is located and a decrease in performance; green indicates a decrease in the result of the metric where it is located and an increase in performance; blue indicates a constant result of the metric where it is located and no change in performance. From the table, we can get that ElasticROS completely improves the latency performance. For CPU usage and power consumption performance, ElasticROS also improves performance in the vast majority of cases. 

We can conclude from the SLAM experiments that ElasticROS is able to perform collaborative computing elastically under different resource conditions, thus achieving computing optimization. It achieves better performance compared to the node-level FogROS.
\subsection{Experiments: Grasping}
\begin{figure}[!ht]
\linespread{1.0}
\centering
	\subfloat[The experimental scenario of grasping with the robotic arm. The robotic arm executes the control program after obtaining the results of the grasp detection. The grasp detection is one of the calculation modules for grasping.]{\includegraphics[width = 0.5\textwidth]{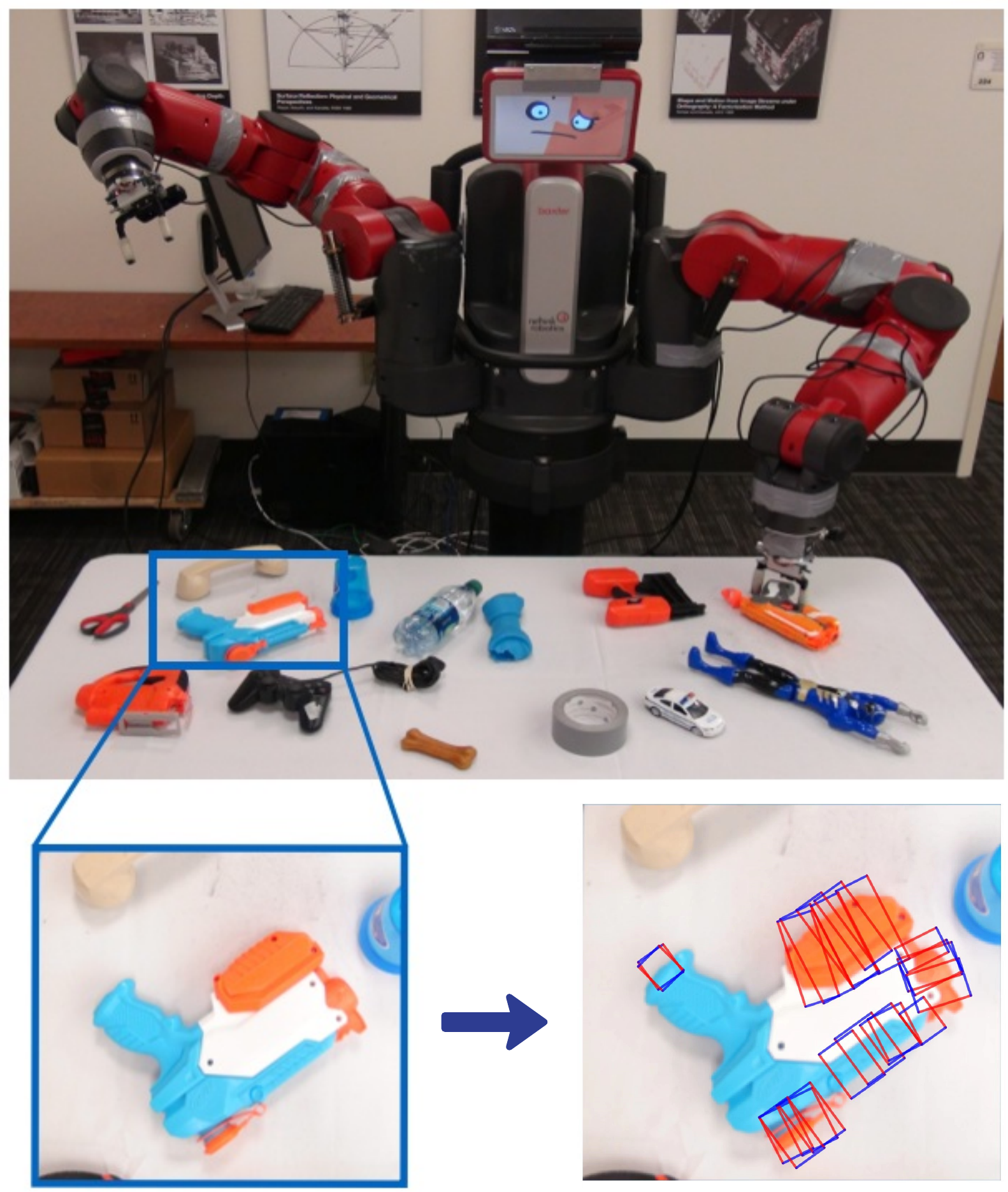}}
	\hfill
	\subfloat[Results of the grasp detection approach.]{\includegraphics[width = 0.5\textwidth]{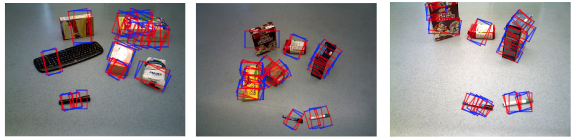}}
	\caption{Scenarios of grasping that can improved with ElasticROS.}
\label{fig:graspScenario}
\end{figure}
Grasping is one of the common tasks of robots, and we perform a collaborative computing upgrade for the grasp detection module present by \cite{chu2018real}. In this experiment, we will test a new challenge, namely bandwidth glitch. Fig. \ref{fig:graspScenario} shows the experimental scenario. The online learning feature of ElasticROS should enable the algorithm to adjust the parameters in response to sudden bandwidth changes to ensure that the system performs elastically in an optimal way.
\begin{figure*}
    \centering
    \includegraphics[width=1\textwidth]{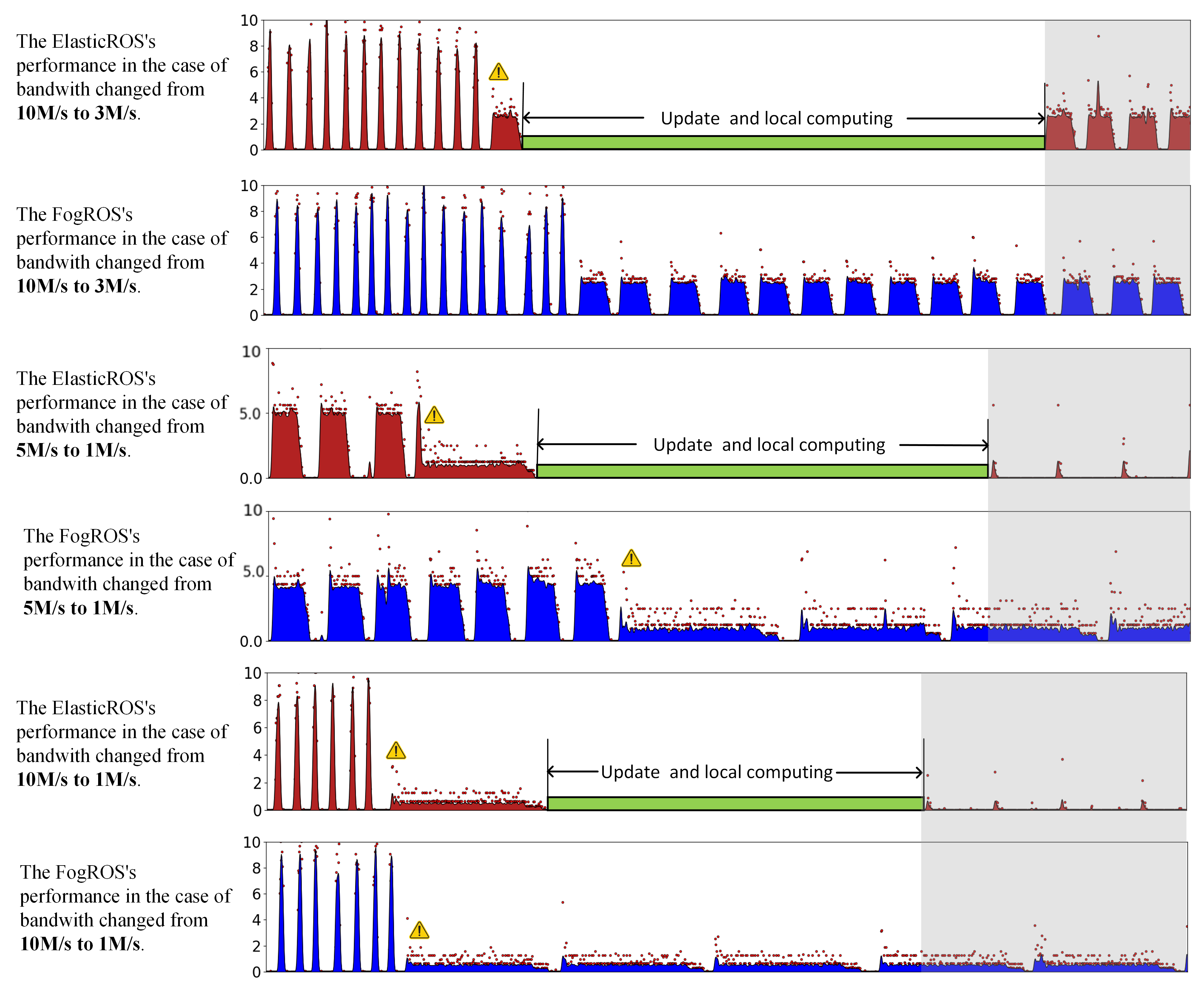}
    \linespread{1.0}
    \caption{Performance comparison of ElasticROS and FogROS in the experiment of robotic grasping task. Green parts indicate parameters updating. Gray shaded parts indicate the collaborative computing strategy after parameter update. The yellow markers denote the sudden change of bandwidth.}
    \label{graspcompare}
\end{figure*}

Fig. \ref{graspcompare} depicts the comparison of the performance of ElasticROS and FogROS under the sudden change in bandwidth, which also reflects the computing frequency and performance. The first two rows present a comparison of computing in the case of a sudden change in bandwidth from 10M/s to 3M/s. The figure in the first row indicates that ElasticROS is able to detect the prediction function anomaly and initiate parameter updating in time when the bandwidth changes abruptly. The parameters are fully cloud computing before and after updating with the same as FogROS. The third and fourth rows present a comparison of the computing in the case of a sudden change in bandwidth from 5M/s to 1M/s. ElasticROS initiates the parameters updating to get a new function in this case and gets a better performance than FogROS. The last two rows compare the computing in the case of a sudden change in bandwidth from 5M/s to 1M/s. ElasticROS achieves a performance improvement of about three times over FogROS.

We can conclude from the grasping experiments that ElasticROS is able to update the parameters and achieve online learning in case of sudden changes in the environment. A better performance is achieved compared to the node-level FogROS.
\linespread{1.5}
\begin{table*}[htbp]
  \centering
  \caption{Comparison of experimental results for FogROS and ElasticROS in the grasping task with sudden bandwidth changes.}
    \resizebox{1\textwidth}{!}{
    \begin{tabular}{ccccccccccc}
    \toprule
    \multicolumn{2}{c}{\multirow{2}[4]{*}{Matrics}} & \multicolumn{3}{c}{10M{\color{black}{$\rightarrow$}}3M} & \multicolumn{3}{c}{5M{\color{black}{$\rightarrow$}}1M} & \multicolumn{3}{c}{10M{\color{black}{$\rightarrow$}}1M} \\
    \cmidrule(lr){3-5} \cmidrule(lr){6-8} \cmidrule(lr){9-11}&  & Before\quad{\color{black}{$\rightarrow$}} & On\quad{\color{black}{$\rightarrow$}} &\textbf{*After}& Before\quad{\color{black}{$\rightarrow$}} & On\quad{\color{black}{$\rightarrow$}}    & \textbf{*After} & Before\quad{\color{black}{$\rightarrow$}} & On\quad{\color{black}{$\rightarrow$}}    &\textbf{*After}\\
    \midrule
    \multirow{3}[2]{*}{FogROS} & Latency & 0.27  & 1.82  & \textbf{1.82}\quad{\color{green}{$\downarrow$}} & 0.27  & 5.21  & \textbf{5.21}\quad{\color{red}{$\uparrow$}} & 0.27  & 5.28  & \textbf{5.28}\quad{\color{red}{$\uparrow$}} \\
 & CPU& 3.10\% & 21\%     & \textbf{21\%}\quad{\color{green}{$\downarrow$}} & 11.70\% & 61\% & \textbf{61\%}\quad{\color{red}{$\uparrow$}} & 3.10\% & 59\%  & \textbf{62\%}\quad{\color{red}{$\uparrow$}} \\
 & Power& 255   & 1718     & \textbf{1718}\quad{\color{green}{$\downarrow$}} & 920   & 4925  & \textbf{4925}\quad{\color{red}{$\uparrow$}} & 257   & 4924  & \textbf{4924}\quad{\color{red}{$\uparrow$}} \\
    \midrule
    \multirow{3}[2]{*}{ElasticROS} & Latency & 0.27  & 2.61  & \textbf{1.82}\quad{\color{green}{$\downarrow$}} & 0.27  & 2.6   & \textbf{1.9} \quad{\color{green}{$\downarrow$}} & 0.27  & 2.61  & \textbf{1.9}\quad{\color{green}{$\downarrow$}} \\
& CPU& 3.10\% & 22\%      & \textbf{22\% }\quad{\color{green}{$\downarrow$}} & 11.70\% & 96\%  & \textbf{58\%}\quad{\color{green}{$\downarrow$}} & 3.10\% & 96\%  & \textbf{59\%}\quad{\color{green}{$\downarrow$}} \\
& Power & 257   & 4682     & \textbf{1752}\quad{\color{green}{$\downarrow$}} & 920   & 4632  & \textbf{3490}\quad{\color{green}{$\downarrow$}} & 257   & 4599  & \textbf{3472}\quad{\color{green}{$\downarrow$}} \\
    \bottomrule
    \end{tabular}%
    }
  \label{tab:graspdata}%
\end{table*}%
\begin{figure*}
    \centering
    \includegraphics[width=0.9\textwidth]{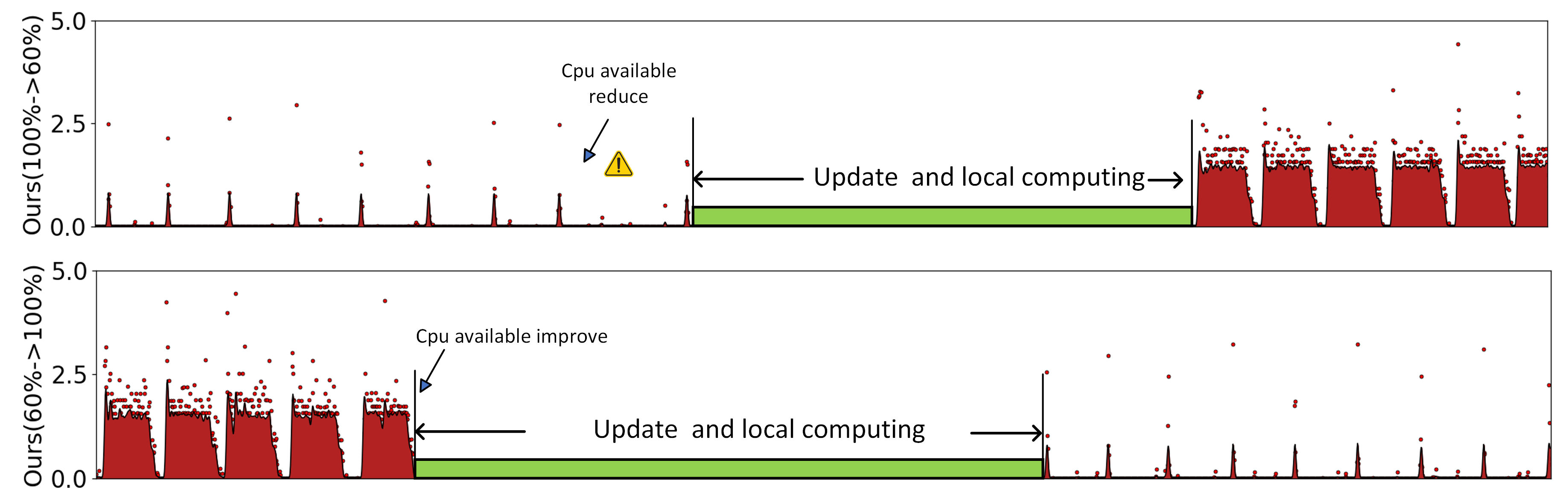}
    \caption{Performance of ElasticROS when CPU availability changes. Green parts indicate parameters updating.}
    \label{cpuchange}
\end{figure*}
Table \ref{tab:graspdata} presents a comparison of the experimental results of FogROS and ElasticROS, including before, on and after the parameter updating. Also, we compare CPU usage and power consumption, noting that these two metrics are calculated relatively that it taking the computing frequency into account. The bolded data in the table shows the performance of ElasticROS after parameter updating, and the arrows on its right side denote the comparison with pure robot computing. The red arrows denote increased results and decreased performance of the metric; the green arrows denote decreased results and increased performance of the metric. The table shows that ElasticROS completely improves the performance of the system.
\subsection{Experiments: Human-robot dialogue}
\begin{figure}
    \centering
    \includegraphics[width=0.45\textwidth]{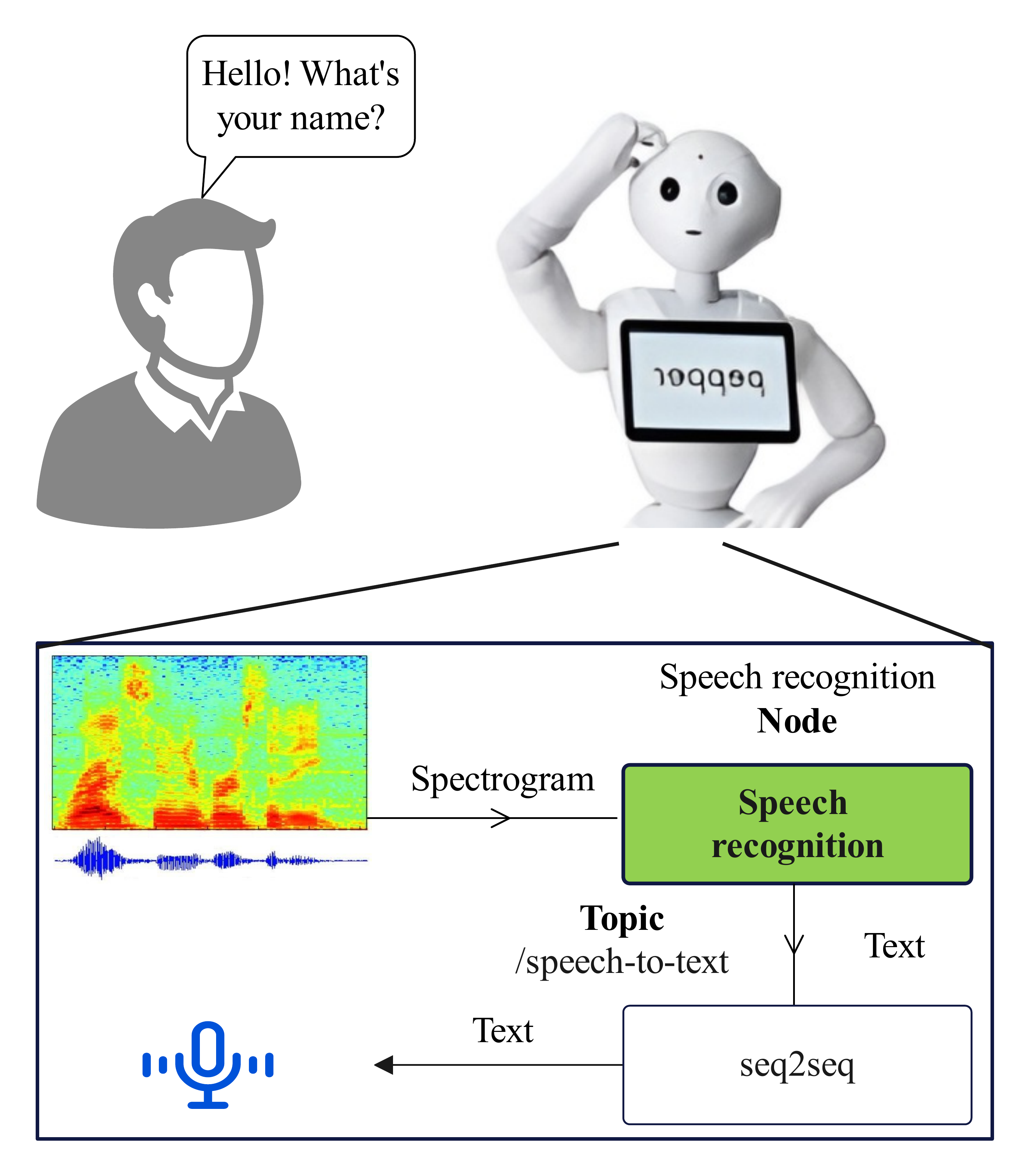}
    \linespread{1.0}
    \caption{The human-robot dialogue scenario and computing processes. This work conducts experiments on speech recognition nodes.}
    \label{humanrobot}
\end{figure}
Robots often have many different functions, and the computing to be carried out at the same time will change. Correspondingly, for a single algorithm, the available CPU is also changing, so it is necessary for us to carry out experiments under the change of available CPU usage. Robot-Human dialogue is an interactive task often performed between robots and people in real life. We take this as an example to verify the performance of ElasticROS in the case of available CPU usage changes. Fig.\ref{humanrobot} illustrates the computing flow of the robot-human dialogue. We performed an ElasticROS deployment on the speech recognition module \cite{beckmann2019speech}. We only performed the experiment on ElasticROS but without FogROS, since FogROS is not correlated with CPU usage.

Fig. \ref{cpuchange} depicts the performance of ElasticROS in this experiment. In the first row, when available CPU usage decreases, ElasticROS updates the parameters and adjusts the policy afterwards, choosing to compute less in the robot but transmit more data. In the second row, when the available CPU usage increases, ElasticROS updates the parameters and adjusts the policy afterwards, choosing a policy that computes more in the robot but transmits less data. From the figure, we can obtain that the elastic node in ElasticROS initiates a parameter update after an inaccuracy in the prediction function and obtains a new collaborative computing strategy. It demonstrates the adaptive capability of ElasticROS to CPU usage changes.
\linespread{1.5}
\begin{table}
  \centering
  \caption{Experimental results for ElasticROS in the human-robot dialogue task with CPU usage changes.}
    \begin{tabular}{cccccc}
    \toprule
    \multicolumn{2}{c}{Matrics} & Before{\color{black}{$\rightarrow$}} & If remain{\color{black}{$\rightarrow$}} & On{\color{black}{$\rightarrow$}}    & After \\
    \cmidrule(lr){1-2} \cmidrule(lr){3-6}\multirow{2}[2]{*}{Latency} & ElasticROS & 1.76s & \textbf{2.67s} & 3.92s & \textbf{1.71s}\ {\color{green}{$\downarrow$}} \\
          & Robot & 2.62s & 3.92s & 3.92s & 3.92s \\
    \midrule
    \multirow{2}[2]{*}{CPU rela.} & ElasticROS & 55\%  & \textbf{54\%}  & 1+46\% & \textbf{13\%}\ {\color{green}{$\downarrow$}} \\
          & Robot & 98\%  & \multicolumn{3}{c}{relative 89\%, absolute 60\%} \\
    \midrule
    \multirow{2}[2]{*}{Power rela.} & ElasticROS & 3067  & \textbf{4689}  & 4689  & \textbf{1631}\ {\color{green}{$\downarrow$}} \\
          & Robot & 4672  & \multicolumn{3}{c}{relative 6990, absolute 4672} \\
    \bottomrule
    \end{tabular}%
  \label{tab:cpu}%
\end{table}%
Table \ref{tab:cpu} shows the experimental results of ElasticROS, from which we can get that ElasticROS completely improves the performance of the system after the parameters are updated compared to FogROS.

\section{Conclusion}
The increasing number of models integrated into robots, the steep increase of model parameters, the theoretical bottleneck of parameters abatement, and the bottleneck of robot battery capacity block the progress of robots into real life. In other words, it is paradoxical that we cannot equip every robot with high-performance graphics cards, coolers, and high-capacity batteries, whereas we are trying to improve performance and reduce costs for them simultaneously. Cloud robotics is the most anticipated theory in the robotics community to explore breaking these bottlenecks. In this work, we present ElasticROS, which evolves the current node-level system into an algorithm-level system. ElasticROS is the first robot operating system with algorithm-level collaborative computing for fog and cloud robotics based on ROS. It is also advanced that it realizes self-adaptability to dynamic conditions in an elastic collaborative mode.

We present the ElasticAction algorithm based on online learning in ElasticROS, which determines the way the robot and the server collaborate in a resilient way. The algorithm dynamically adjusts parameters to adapt to changes in the conditions the robot is currently under. Furthermore, we prove that the upper bound of regret in the algorithm is sublinear, which guarantees its convergence and thus makes ElasticROS elastic and stable. Finally, we validate ElasticROS with SLAM, grasping, and human-robot dialogue tasks, and then measure its performance in terms of latency, CPU usage, and power consumption. ElasticROS significantly outperforms baseline and current approaches.

Despite the promising results, we only considered the single-robot scenario and did not analyze the scenario where multi-robots compete for resources. Future work should, therefore, includes more robots and optimizes the overall multi-robot system instead of one robot.Overall, we expect that cloud robotics will be well applied to all areas of robotics. With the development of 6G communication technology, cloud robotics will dominate the future of robotics computing.

\bibliographystyle{IEEEtran}
\bibliography{sample}
\end{document}